\documentclass{article} 
\usepackage{iclr2024_conference,times}

\usepackage{amsmath,amsfonts,bm}

\def\eqref#1{equation~\ref{#1}}

\def\1{\bm{1}}

\DeclareMathAlphabet{\mathsfit}{\encodingdefault}{\sfdefault}{m}{sl}
\SetMathAlphabet{\mathsfit}{bold}{\encodingdefault}{\sfdefault}{bx}{n}

\usepackage{hyperref}
\usepackage{url}
\usepackage{makecell}
\usepackage{graphicx}
\usepackage{subfig}
\usepackage{float}
\usepackage{multirow}
\usepackage{booktabs}
\usepackage{wrapfig}
\usepackage{verbatim}

\title{OccuQuest: Mitigating Occupational Bias for Inclusive Large Language Models}

\author{Mingfeng Xue\thanks{ Work done during internships at Alibaba Group.} , \:\:\:
\textbf{Dayiheng Liu}\thanks{Corresponding author} \textbf{,} \:\:\:
\textbf{Kexin Yang}$^\ast$\textbf{,} \:\:\:
\textbf{Guanting Dong}$^\ast$\textbf{,} \:\:\:
\textbf{Wenqiang Lei,} \:\:\: \\
\textbf{Zheng Yuan,} \:\:\:
\textbf{Chang Zhou,} \:\:\:
\textbf{Jingren Zhou} \\
Alibaba Group \\
\texttt{\{xuemingfeng.xmf, liudayiheng.ldyh\}@alibaba-inc.com}
}

\iclrfinalcopy 
\begin{document}

\maketitle
\begin{abstract}
The emergence of large language models (LLMs) has revolutionized natural language processing tasks.
However, existing instruction-tuning datasets suffer from occupational bias: the majority of data relates to only a few occupations, which hampers the instruction-tuned LLMs to generate helpful responses to professional queries from practitioners in specific fields.
To mitigate this issue and promote occupation-inclusive LLMs, we create an instruction-tuning dataset named \emph{OccuQuest}, which contains 110,000+ prompt-completion pairs and 30,000+ dialogues covering over 1,000 occupations in 26 occupational categories.
We systematically request ChatGPT, organizing queries hierarchically based on Occupation, Responsibility, Topic, and Question, to ensure a comprehensive coverage of occupational specialty inquiries.
By comparing with three commonly used datasets (Dolly, ShareGPT, and WizardLM), we observe that OccuQuest exhibits a more balanced distribution across occupations.
Furthermore, we assemble three test sets for comprehensive evaluation, an occu-test set covering 25 occupational categories, an estate set focusing on real estate, and an occu-quora set containing real-world questions from Quora.
We then fine-tune LLaMA on OccuQuest to obtain OccuLLaMA, which significantly outperforms state-of-the-art LLaMA variants (Vicuna, Tulu, and WizardLM) on professional questions in GPT-4 and human evaluations.
Notably, on the occu-quora set, OccuLLaMA reaches a high win rate of 86.4\% against WizardLM.
Furthermore, we demonstrate the potential of combining OccuQuest with other instruction-tuning datasets to enhance the overall performance of LLMs.
By fine-tuning LLaMA on a mixture of OccuQuest and Tulu datasets, we introduce ProLLaMA, which excels in addressing occupational questions and exhibits superior performance in comprehensive evaluations such as MMLU, GSM8K, BBH, and HumanEval.
Among the different LLaMA variants, the 7B and 13B ProLLaMA models achieve the highest performance on MMLU and GSM8K, with the 7B ProLLaMA model demonstrating an improvement of more than 4 points over the other 7B variants on GSM8K.
We open release the dataset and models.\footnote{OccuQuest: \url{https://huggingface.co/datasets/OFA-Sys/OccuQuest}, OccuLLaMA: \url{https://huggingface.co/OFA-Sys/OccuLLaMA-7B}, ProLLaMA: \url{https://huggingface.co/OFA-Sys/ProLLaMA-7B}.}
\end{abstract}
\section{Introduction}
\label{sec_intro}
The emergence of large language models (LLMs), such as GPT~\citep{DBLP:conf/nips/BrownMRSKDNSSAA20, DBLP:conf/nips/Ouyang0JAWMZASR22, DBLP:journals/corr/abs-2303-08774}, PaLM~\citep{DBLP:journals/corr/abs-2204-02311, DBLP:journals/corr/abs-2210-11416}, LLaMA~\citep{DBLP:journals/corr/abs-2302-13971} and its various variants, trigger a paradigm shift in natural language processing (NLP) tasks.
Instruction tuning has become a crucial process following pre-training, aiming to align the behavior of LLMs with human expectations.
However, we observe a notable \textbf{occupational bias} in the existing instruction-tuning datasets, as a significant portion of data is centered around specific occupational groups.
Unfortunately, language models tend to capture and reflect this bias~\citep{suresh2021understanding, DBLP:conf/naacl/ShenHCBF22, DBLP:conf/acl/Lee0PKKH23}, making it challenging for them to generate accurate and insightful responses to questions from specific occupations.
\begin{figure*}[t]
    \centering
    \subfloat[Dolly]{
        \label{fig_dataset_category_dolly}\includegraphics[width=0.48\textwidth]{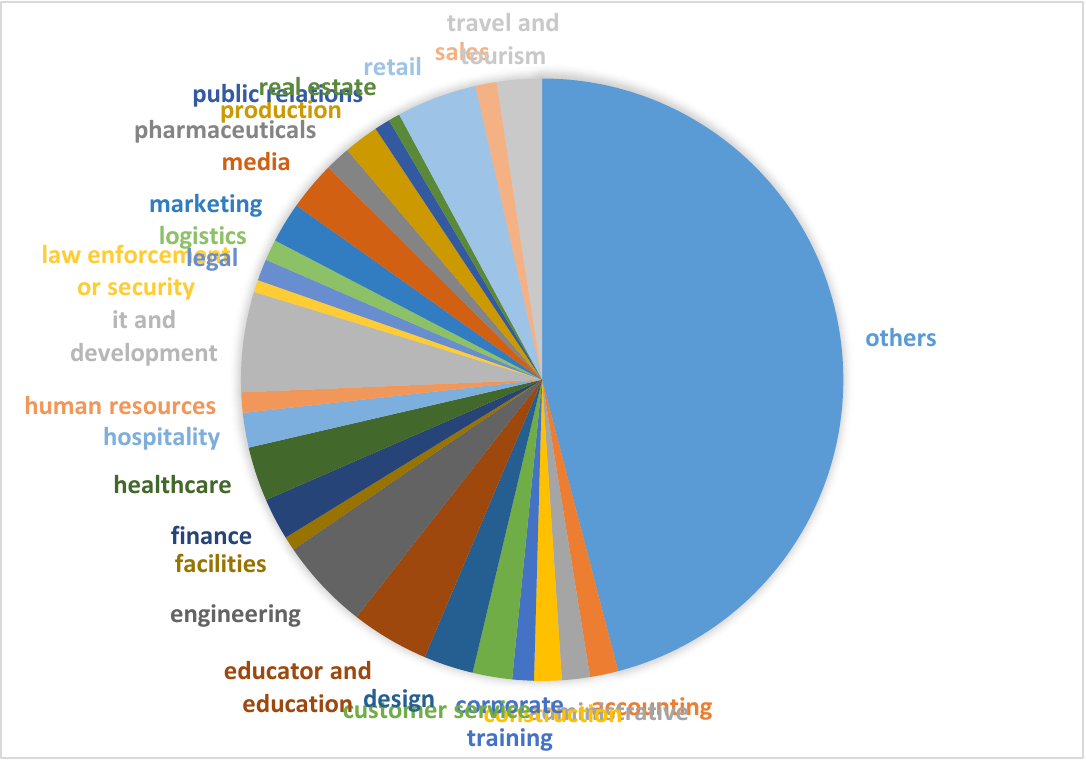}
        }
    \subfloat[ShareGPT]{
        \label{fig_dataset_category_sharegpt}\includegraphics[width=0.48\textwidth]{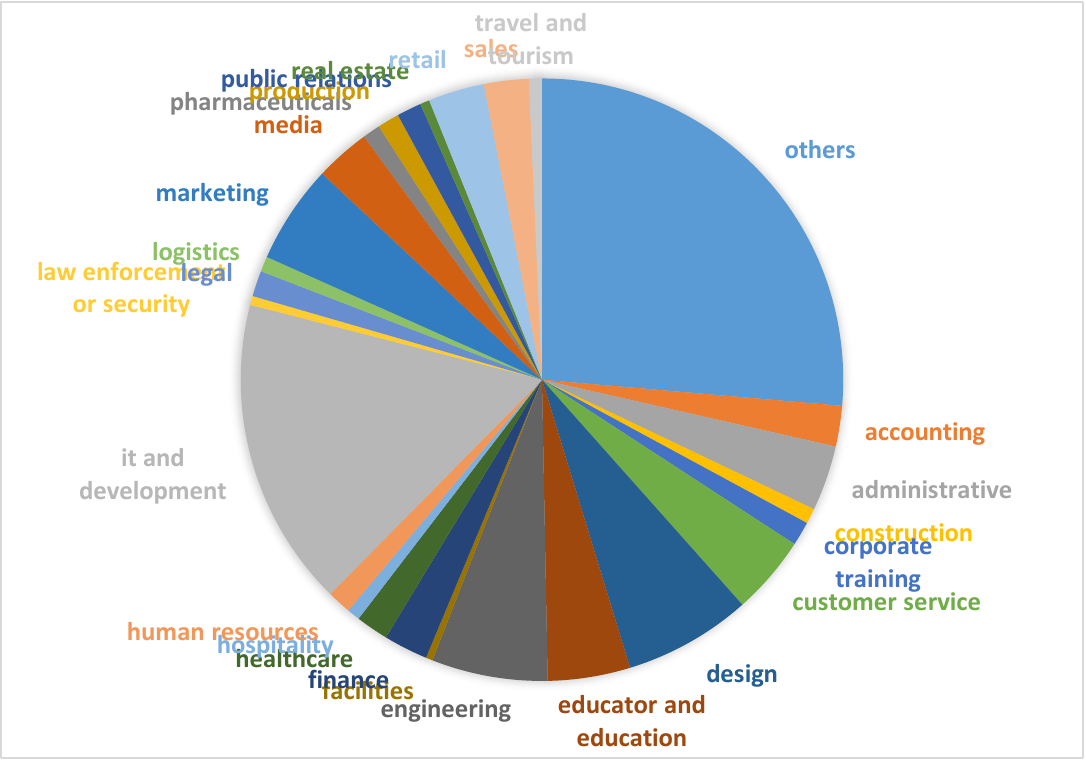}
        }
    \\
    \subfloat[WizardLM]{
        \label{fig_dataset_category_wizardlm}\includegraphics[width=0.48\textwidth]{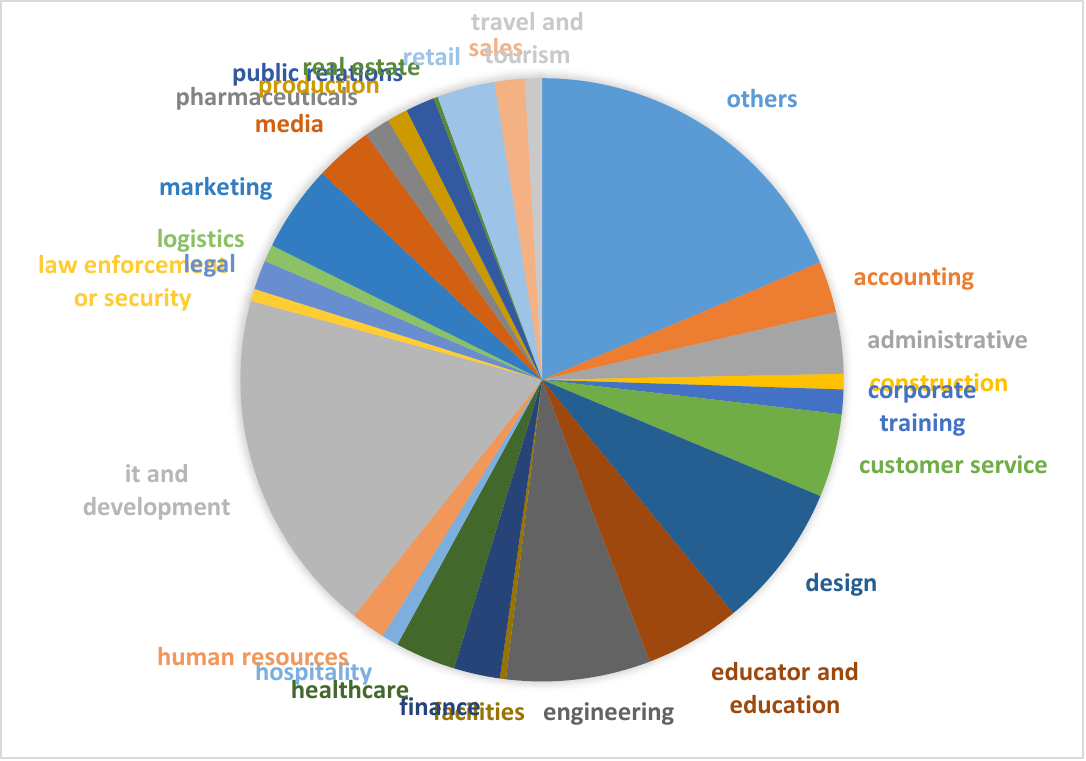}
        }
    \subfloat[OccuQuest]{
        \label{fig_dataset_category_occuquest}\includegraphics[width=0.48\textwidth]{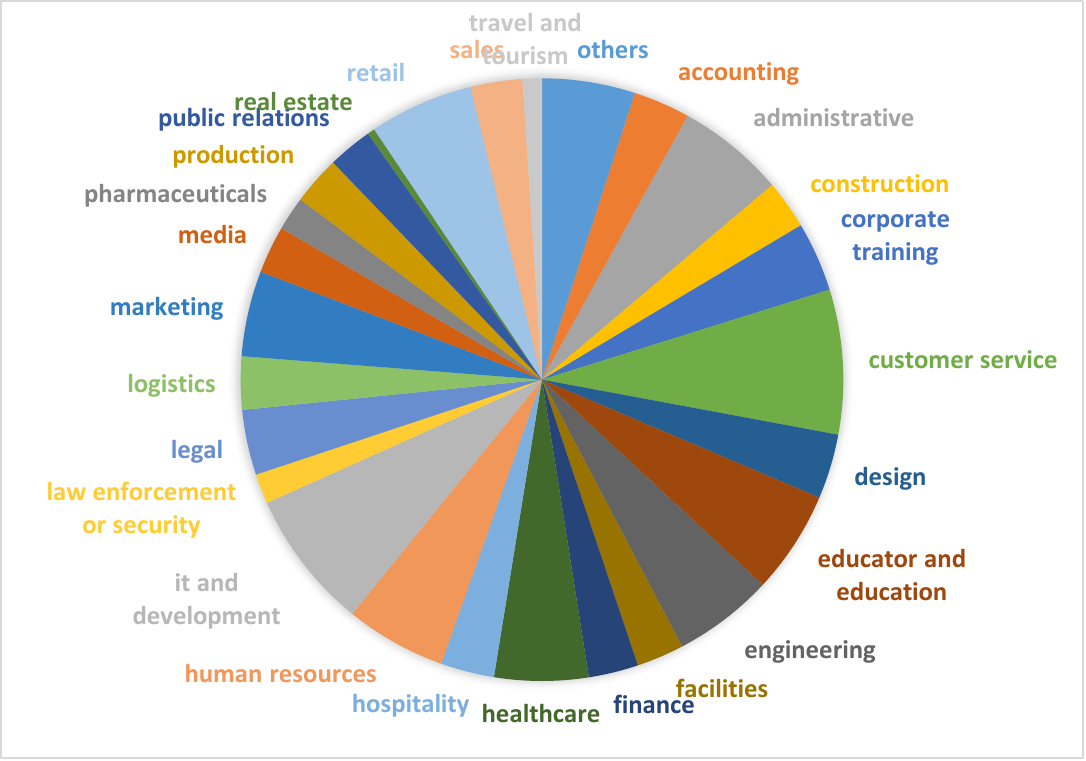}
        }
    \caption{The distribution of occupational categories across various datasets.}
    \label{fig_dataset_category}
\end{figure*}
\par
The primary origins of instruction-tuning data encompass pre-existing NLP tasks (Flan~\citep{DBLP:conf/iclr/WeiBZGYLDDL22}, SuperNI~\citep{DBLP:conf/emnlp/WangMAKMNADASPK22}), manually constructed instructions (Dolly~\citep{DatabricksBlog2023DollyV2}, OpenAssistant~\citep{DBLP:journals/corr/abs-2304-07327}), and datasets generated using LLMs (Alpaca~\citep{alpaca}, ShareGPT\footnote{\url{https://sharegpt.com/}}, WizardLM~\citep{DBLP:journals/corr/abs-2304-12244}).
Practitioners in the AI industries are more likely to access these sources, as Databricks admits that the Dolly dataset comes from over 5,000 employees who are very interested in LLMs~\citep{DatabricksBlog2023DollyV2}.
However, practitioners across various fields with weak connections to AI communities have limited access to these data sources.
A welder stands less chance of producing instruction-tuning data than an employee of an AI institute.
Consequently, while the LLMs fine-tuned on these datasets excel in answering queries related to building chatbots, they may struggle with questions about rectifying the lack of fusion in welding.
These allocative harms~\citep{barocas2017problem} hamper the LLMs from providing helpful, honest, and harmless~\citep{DBLP:journals/corr/abs-2112-00861} assistance to specific occupational groups.
\par
To create more inclusive and unbiased language models that can better serve users from different occupational backgrounds, we propose an instruction-tuning dataset that covers over 1,000 occupations.
We collect over 1,000 job titles and their responsibilities spanning 26 distinct occupation categories from Workable\footnote{\url{https://resources.workable.com/job-descriptions/}}.
We illustrate the categories and the representative occupations in Appendix~\ref{lab_occupational_categories}.
We then utilize ChatGPT\footnote{In the study, we use the gpt-3.5-turbo API (\url{https://platform.openai.com/docs/models/gpt-3-5}) because it is capable and fee friendly.} to identify key topics of concern for practitioners in each occupation and generate relevant questions and answers accordingly.
This effort results in the creation of a comprehensive dataset called OccuQuest comprising 148,772 queries and responses, covering 1,013 occupations and 31,811 topics.
\par
We compare the distribution of occupational categories in OccuQuest with three typical instruction-tuning datasets (Dolly, ShareGPT, and WizardLM) using ChatGPT, and Figure~\ref{fig_dataset_category} illustrates the results.
The precise distribution percentages are presented in Appendix~\ref{appendix_dataset_statistics}.
Our analysis reveals that Dolly, ShareGPT, and WizardLM favor non-occupation-related topics (denoted as \emph{Others}) and the "IT and Development" category, while OccuQuest exhibits a more balanced distribution.
For instance, ShareGPT and WizardLM consist of less than 0.8\% of data in the "Facilities" category, while comprising over 15\% of data in "IT and Development".
Conversely, in OccuQuest, the majority of occupational categories encompass data ranging from 2\% to 6\%.
\par
To validate the effectiveness of OccuQuest, we fine-tune LLaMA on OccuQuest to get OccuLLaMA and compare it with the state-of-the-art LLaMA variants (Vicuna~\citep{vicuna2023}, WizardLM, and Tulu~\citep{DBLP:journals/corr/abs-2306-04751}) through preference assessments using GPT-4\footnote{\url{https://platform.openai.com/docs/models/gpt-4}} and human evaluations.
OccuLLaMA consistently outperforms other variants in answering occupational questions across various occupations.
Notably, on a test set consisting of real-world questions covering 25 occupational categories, OccuLLaMA achieves an 86.4\% win rate against WizardLM.
\par
Moreover, we demonstrate that the OccuQuest dataset can be effectively combined with other instruction-tuning datasets to enhance the comprehensive abilities of LLMs.
Following \citet{DBLP:journals/corr/abs-2306-04751}, we fine-tune LLaMA on a mixture of OccuQuest and Tulu datasets to obtain ProLLaMA.
ProLLaMA excels in addressing occupational questions and performs well in comprehensive ability evaluations such as MMLU~\citep{DBLP:conf/iclr/HendrycksBBZMSS21}, GSM8K~\citep{DBLP:journals/corr/abs-2110-14168}, BBH~\citep{DBLP:conf/acl/SuzgunSSGTCCLCZ23}, and HumanEval~\citep{DBLP:journals/corr/abs-2107-03374}.
When compared to the above LLaMA variants, the 7B and 13B ProLLaMA models achieve the best performance on MMLU and GSM8K.
In particular, on GSM8K, the 7B ProLLaMA surpasses these 7B variants by a margin exceeding 4 points.
\par
In summary, this article makes four main contributions:
\begin{enumerate}
  \item We propose the OccuQuest dataset, which consists of 148,772 queries and responses covering 1,013 occupations. To the best of our knowledge, this is the first dataset available that focuses on mitigating the issue of occupational bias in LLMs.
  \item We demonstrate the effectiveness of OccuQuest through preference tests with GPT-4 and human evaluations. Additionally, we showcase the integration of OccuQuest with existing datasets to enhance LLMs in a synthetic manner.
  \item We propose ProLLaMA, a series of LLaMA models that excel in answering questions from different occupations and perform well on the comprehensive abilities assessments.
  \item We openly release our dataset and model parameters, encouraging further research and exploration in this domain.
\end{enumerate}
\section{Related Works}
\subsection{Bias in Datasets}
The utilization of deep neural networks relies heavily on datasets, yet existing datasets contain a wide variety of biases including race~\citep{DBLP:conf/naacl/ManziniLBT19, DBLP:conf/fat/SambasivanAHDP21, DBLP:conf/acl/Lee0PKKH23, DBLP:conf/fat/FieldCGCPST23}, gender~\citep{DBLP:conf/ethnlp/KoolenC17, DBLP:conf/naacl/RudingerNLD18}, disability~\citep{DBLP:conf/acl/HutchinsonPDWZD20, DBLP:conf/fat/GadirajuKDTWDB23}, and others.
Extensive researches uncover and analyze these biases in traditional NLP tasks~\citep{DBLP:journals/corr/abs-1909-05088, DBLP:conf/aies/0002SAKFLP18}.
Additionally, there is a growing recognition of the social implications and consequences of these biases~\citep{DBLP:conf/acl/HovyS16, barocas2017problem}.
\par
One prominent and effective approach to address biases involves constructing or transforming the datasets.
For instance, \citet{costa2020fine} and \citet{DBLP:conf/acl/SaundersB20} fine-tune models on carefully screened and balanced data to mitigate biases.
\citet{DBLP:conf/acl/WangRC22} adopt data augmentation strategies by randomly switching entities to prevent the translation system from associating specific names with contextual idiosyncrasies.
\citet{DBLP:conf/emnlp/ChoubeyCMD21} generate gender-specific pseudo-parallel corpora to prompt translation systems to produce accurate gender-specific translations.
Motivated by the insights from these studies, we aim to mitigate occupational bias in LLMs by constructing an occupationally balanced instruction-tuning dataset.
To the best of our knowledge, this is the first endeavor specifically targeting the mitigation of occupational bias in LLMs.
\subsection{Instruction Tuning}
In recent years, there has been significant progress in LLMs, with notable advancements demonstrated by GPT-3~\citep{DBLP:conf/nips/BrownMRSKDNSSAA20}, highlighting the potential of context learning in LLMs.
As a result, numerous LLMs have emerged, such as Gopher~\citep{DBLP:journals/corr/abs-2112-11446}, Chinchilla~\citep{DBLP:journals/corr/abs-2203-15556}, and PaLM~\citep{DBLP:journals/corr/abs-2204-02311}, showcasing exceptional performance and dominance across diverse NLP tasks.
\par
To align the behavior of LLMs with human preferences, instruction tuning has emerged as a crucial method~\citep{DBLP:conf/nips/Ouyang0JAWMZASR22, DBLP:journals/corr/abs-2210-11416}.
There are three primary sources of existing instruction-tuning datasets: datasets derived from pre-existing NLP tasks, datasets created through manual authoring, and datasets generated using LLMs.
Initially, instruction-tuning datasets are developed by expanding upon existing NLP task datasets.
For example, Flan~\citep{DBLP:conf/iclr/WeiBZGYLDDL22} and SuperNI~\citep{DBLP:conf/emnlp/WangMAKMNADASPK22} are designed by converting data from diverse NLP tasks, such as classification, extraction, and infilling, into instructions, inputs, and outputs format using templates.
However, using existing datasets has the drawback of limited diversity in topics and syntax within the instructions.
To overcome this limitation, Dolly~\citep{DatabricksBlog2023DollyV2} and OpenAssistant~\citep{DBLP:journals/corr/abs-2304-07327} enhance diversity by manually crafting prompts.
Additionally, recent studies have explored cost-effective and efficient approaches by leveraging ChatGPT to obtain prompts and responses, reducing the costs and labor involved~\citep{DBLP:conf/acl/WangKMLSKH23, alpaca, DBLP:journals/corr/abs-2304-12244, vicuna2023, DBLP:journals/corr/abs-2304-01196}.
\par
Extensive efforts have been dedicated to augmenting instruction-tuning datasets, aiming to improve the generalization capabilities of LLMs.
However, these existing datasets exhibit a limited occupational distribution, resulting in inadequate precision and granularity of responses to queries from specific occupations.
This study aims to construct an instruction-tuning dataset that encompasses a wide range of occupation-related topics, thereby mitigating the occupational bias present in LLMs.
\section{OccuQuest Dataset}
\subsection{Dataset Construction}
To mitigate the issue of occupational bias in the instruction-tuning corpus, we intend to construct a dataset that encompasses a wide range of occupational specializations.
We request ChatGPT hierarchically, focusing on Occupation, Responsibility, Topic, and Questions, to cover as many occupations and their corresponding areas of interest as possible.
The data construction process consists of five steps, which are outlined below.
\par
\textbf{Step 1: get occupations.}
To begin, we gather occupation titles and their associated responsibilities.
Workable offers more than 1,000 occupational titles organized into 26 occupational categories.
Each occupation is accompanied by a list of responsibilities, consisting of one sentence per responsibility.
We successfully collect 1,037 occupations and their respective responsibilities from Workable, with an average of around 7 responsibilities per occupation.
\par
\textbf{Step 2: request topics.}
We utilize ChatGPT to generate multiple related topics and topic features by providing the occupation name and one responsibility.
A topic is a keyword or keywords that reflect what a practitioner needs to consider when fulfilling a specific responsibility and topic features provide a descriptive paragraph about the topic.
To avoid duplication, we employ MinHash~\citep{DBLP:conf/cpm/Broder00} on topic features to filter out topics that exhibit high similarities.
\par
\textbf{Step 3: request prompts.}
Using the topic and topic features obtained in Step 2, we request ChatGPT to generate multiple prompts describing potential queries that practitioners may encounter.
During the request, ChatGPT is asked to list the keywords and then generate the prompts to produce diverse prompts with distinct keywords, as directly generating prompts tends to result in similar prompts.
Same as the topic filtering process, we filter out the prompts that show high similarities.
\par
\textbf{Step 4: get responses.}
In this step, we ask ChatGPT to answer the prompts generated in Step 3.
To improve the accuracy of the completions, we assign ChatGPT a role corresponding to the occupation before some of the queries.
We also remove responses that contain overly similar completions.
\par
\textbf{Step 5: create dialogs.}
After completing Step 4, we have data for a single round of queries and responses in the OccuQuest dataset.
To enhance the model's ability to handle multi-round requests, we additionally request ChatGPT to generate multi-round dialogues between a rookie and a veteran, discussing problem-solving scenarios encountered at work for each topic.
\par
During Steps 2, 3, 4, and 5, we exclude responses that are less than 50 words in length or contain the phrase "Sorry, as an AI assistant..." to ensure the validity of the responses.
We incur an approximate cost of \$300 for API access fees in the dataset construction process.
\par
\begin{figure*}[t]
    \centering
    \includegraphics[width=\linewidth]{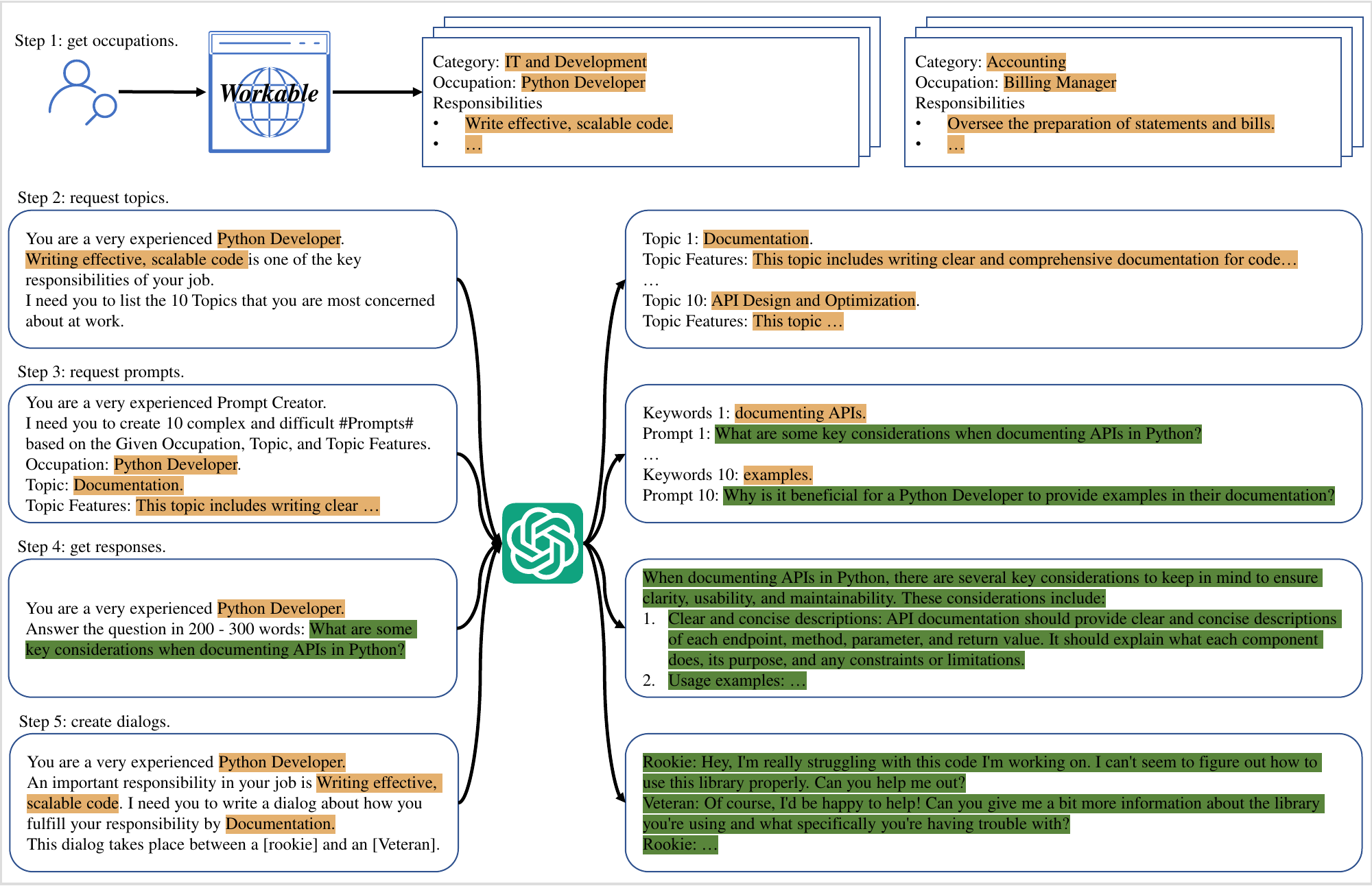}
    \caption{An illustration of the OccuQuest dataset construction process, where the contents highlighted with a background color are ultimately gathered to constitute the dataset. To eliminate duplicate samples, MinHash filtering is applied after steps 2, 3, and 4.}
    \label{fig_dataset_construction}
\end{figure*}

Figure~\ref{fig_dataset_construction} illustrates an example of the dataset construction process, while the actual prompts used to collect the dataset can be found in Appendix~\ref{appendix_whole_prompts}.
The examples extracted from OccuQuest are provided in Appendix~\ref{appendix_examples_in_occuquest}.
\subsection{Dataset Split}
\label{sec_dataset_split}
To assess the efficacy of OccuQuest and the models, we partition a portion of the data from OccuQuest as test sets.
Specifically, we designate the data within the "Real estate" category as the holdout set.
From this category, we randomly select 250 samples, referred to as the "estate set", to evaluate the models' generalization capabilities.
In the remaining 25 categories, we randomly select 100 samples and 10 samples from each as the validation set and "occu-test" set, respectively.
The remaining data in these 25 categories are allocated for the training set.
To ensure the evaluation aligns closely with real-world scenarios, we collect 250 authentic questions (10 questions per category, totaling 25 categories) from Quora\footnote{\url{https://www.quora.com/}} as the "occu-quora" set.
\par
In summary, OccuQuest consists of the following components:
\begin{enumerate}
  \item A training set, containing 114,090 prompt-completion pairs and 31,682 dialogues across 25 categories;
  \item A validation set, containing 2,500 prompt-completion pairs across 25 categories;
  \item An occu-test set, containing 250 prompt-completion pairs across 25 categories;
  \item An estate set, containing 250 prompt-completion pairs in the "Real estate" category;
  \item An occu-quora set, containing 250 real-world questions gathered from Quora across 25 categories.
\end{enumerate}
\subsection{Balanced Distribution of Occupations}
OccuQuest contains 117,090 prompt-completion pairs and 31,682 multi-round dialogues, encompassing 1,013 occupations under 26 occupational categories.
Each item in OccuQuest contains the occupational category, occupation name, topic, topic features, queries, and responses.
Compared to the existing instruction-tuning datasets, OccuQuest exhibits a balanced distribution of occupations.
\par
To evaluate the distribution of occupations within OccuQuest compared to existing instruction-tuning datasets, we select three prominent datasets for comparison: Dolly, ShareGPT, and WizardLM.
Dolly is manually authored, ShareGPT is obtained by the users interacting with ChatGPT, and WizardLM is generated by expanding existing instructions using ChatGPT.
These datasets represent the primary sources of current instruction-tuning data.
We randomly select 10,000 samples from each of the datasets and inquire ChatGPT about the occupational category to which each sample is likely to belong.
The specific prompt used for this task can be found in Appendix~\ref{appendix_whole_prompts}.
\par
The results are presented in Figure~\ref{fig_dataset_category}.
In Dolly, ShareGPT, and WizardLM, the "Others" category unrelated to specific occupations dominates the distribution.
Furthermore, the categories of "IT and Development" and "Engineering" also exhibit a disproportionately high proportion compared to other occupations, consistent with our claim in Section~\ref{sec_intro} that individuals from these fields are more likely to contribute data.
In contrast, OccuQuest demonstrates a more balanced distribution of occupational categories, without any single category displaying clear dominance.
For detailed percentages of occupation distribution across different datasets, please refer to Appendix~\ref{appendix_dataset_statistics}.
\section{Experiments}
\subsection{Baselines}
We fine-tune the LLaMA-7B model on OccuQuest and compare it to competitive baselines:
\par
\textbf{Vicuna}, an open-source chatbot trained by fine-tuning LLaMA on ShareGPT.
Preliminary GPT-4 evaluation reveals that Vicuna-13B achieves over 90\% quality compared to OpenAI ChatGPT~\citep{vicuna2023}.
We utilize the checkpoint available on the Huggingface model repository\footnote{\url{https://huggingface.co/lmsys/vicuna-7b-v1.3}}.
\par
\textbf{Tulu}, a fine-tuned LLaMA on a combination of existing instruction-tuning datasets, including FLAN, Dolly, OpenAssistant, Alpaca, and ShareGPT, proposed by \citet{DBLP:journals/corr/abs-2306-04751}.
Tulu achieves the best average performance on several benchmarks including MMLU, GSM8K, BBH, etc.
We utilize the checkpoint available on the Huggingface model repository\footnote{\url{https://huggingface.co/allenai/tulu-7b}}.
\par
\textbf{WizardLM}, a LLaMA fine-tuned with complex instructions derived from extending the seed instructions in the Alpaca dataset using ChatGPT.
WizardLM achieves more than 90\% capacity of ChatGPT on 17 out of 29 evaluated skills~\citep{DBLP:journals/corr/abs-2304-12244}.
We utilize the checkpoint available on the Huggingface model repository\footnote{\url{https://huggingface.co/WizardLM/WizardLM-7B-V1.0}}.
\par
\textbf{ChatGPT}, a chatbot proposed by OpenAI, recognized as one of the most powerful LLMs (services) currently available.
We use the gpt-3.5-turbo API\footnote{\url{https://platform.openai.com/docs/models/gpt-3-5}} provided by OpenAI for our experiments.
\subsection{Training Details}
To obtain the OccuLLaMA model, we fine-tune the LLaMA-7B model using the OccuQuest training set.
The fine-tuning process involves training for 5 epochs, with a batch size of 128 and a total of 5,500 training steps.
We employ the AdamW optimizer with a maximum learning rate of $2 \times 10^{-5}$, and the learning rate is linearly decayed during training.
Additionally, we set the warmup ratio to 0.03 to gradually increase the learning rate at the beginning of training.
The entire training process is executed on a server equipped with 8 $\times$ 80G A100 GPUs and completes within 8 hours.
\subsection{Evaluation Setup}
We generate the responses to the queries in the occu-test, estate, and occu-quora sets employing the baselines and OccuLLaMA through greedy search, with the maximum generation length set to 1024 tokens.
Subsequently, we evaluate the responses using GPT-4 and human evaluations.
\subsubsection{GPT-4 Evaluation}
The evaluation of open-ended generation using LLMs highlights the benefits of scalability and explainability, and previous studies have shown that GPT-4 exhibits high agreement with human experts~\citep{DBLP:journals/corr/abs-2306-05685}.
Therefore, we leverage GPT-4\footnote{We use the "gpt-4" API in \url{https://platform.openai.com/docs/models/gpt-4}.} to evaluate the performance of OccuLLaMA and the baselines in addressing occupation-related queries.
During this evaluation, we compare the responses from OccuLLaMA with those generated by each baseline.
To ensure fairness and avoid any positional bias, we judge each query twice by swapping the order of the two responses and only declare a win when a response is preferred in both orderings~\citep{DBLP:journals/corr/abs-2306-05685}.
For the specific prompt utilized in the evaluation, please refer to Appendix~\ref{appendix_whole_prompts}.
\subsubsection{Human Evaluation}
We conduct a human evaluation to assess the alignment of the generated responses with human expectations.
Due to the substantial labor costs associated with human evaluation, we randomly select two questions from each occupational category in the occu-test and occu-quora sets, resulting in a human evaluation set comprising 100 samples.
We engage three annotators who are tasked with rating the responses on three dimensions: \textbf{Helpfulness}, \textbf{Honesty}, and \textbf{Harmlessness}~\citep{DBLP:journals/corr/abs-2112-00861}.
These dimensions are assessed on a scale of 1 to 5, with higher scores indicating superior performance.
For more details about the human evaluation, please refer to Appendix~\ref{appendix_human_evaluation_details}.
\subsection{Experimental Results}
\label{sec_experimental_results}
\begin{figure*}[ht]
    \centering
    \subfloat[occu-test]{
        \label{fig_winrate_gpt4_occullama_test}\includegraphics[width=0.321\textwidth]{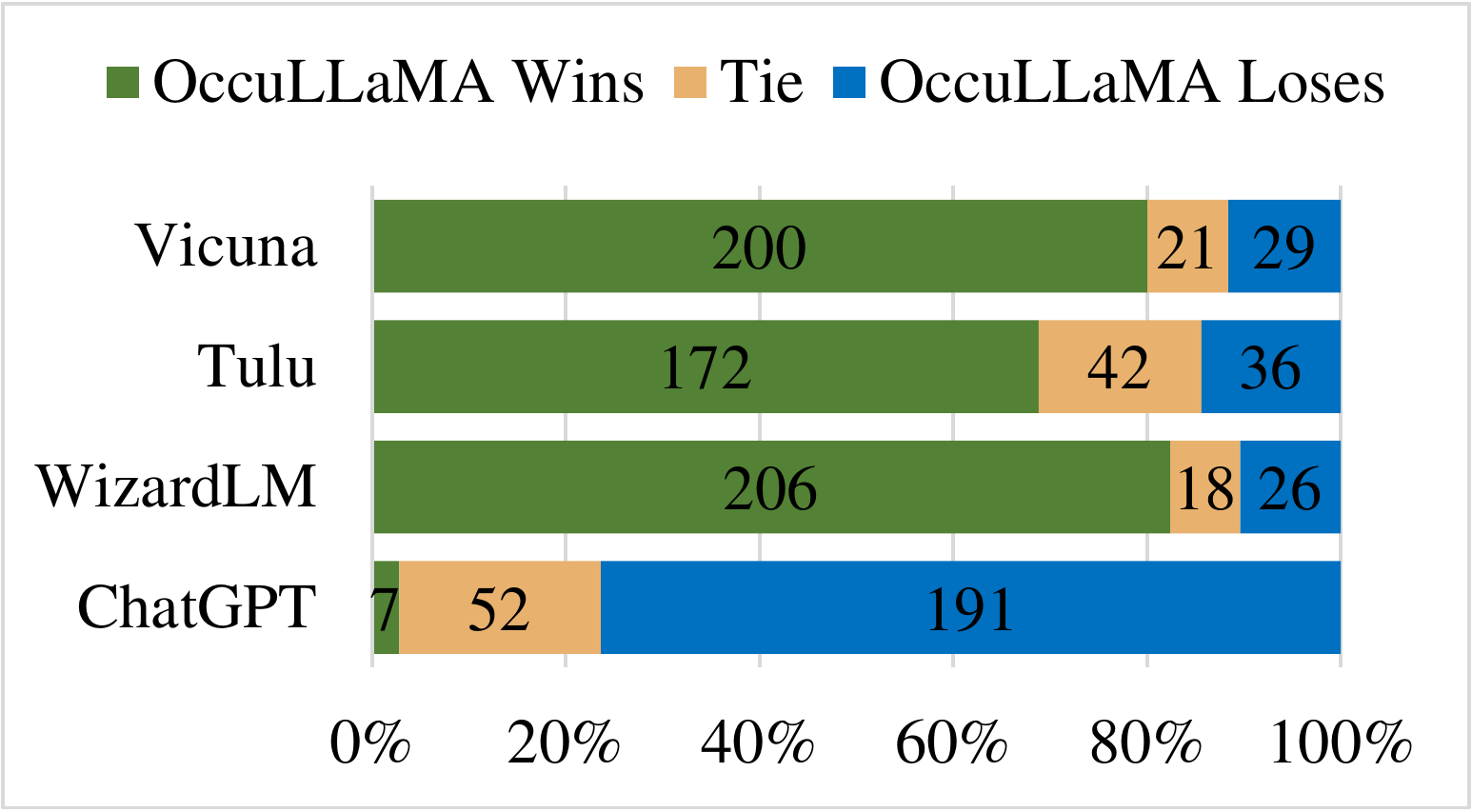}
        }
    \subfloat[estate]{
        \label{fig_winrate_gpt4_occullama_estate}\includegraphics[width=0.321\textwidth]{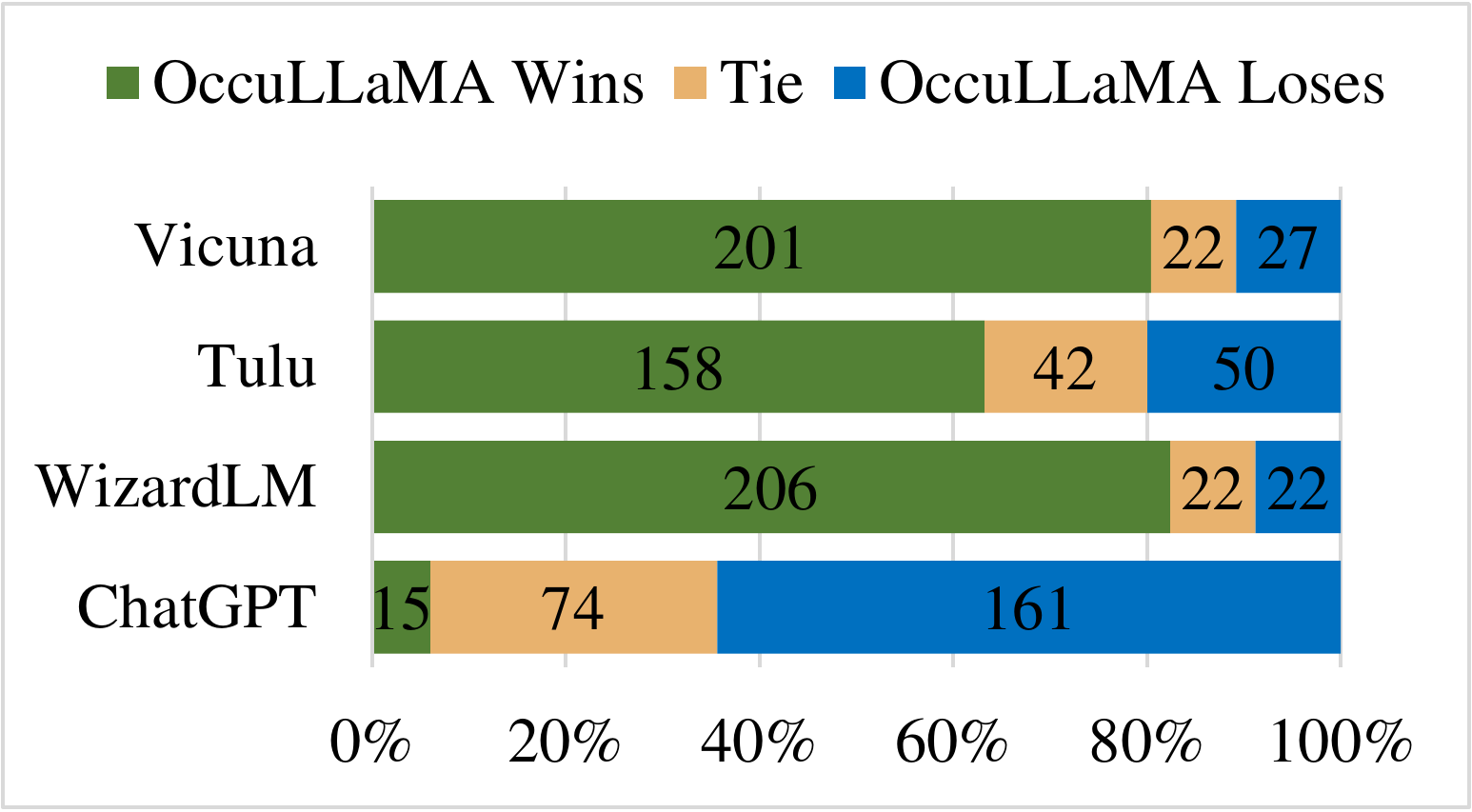}
        }
    \subfloat[occu-quora]{
        \label{fig_winrate_gpt4_occullama_quora}\includegraphics[width=0.321\textwidth]{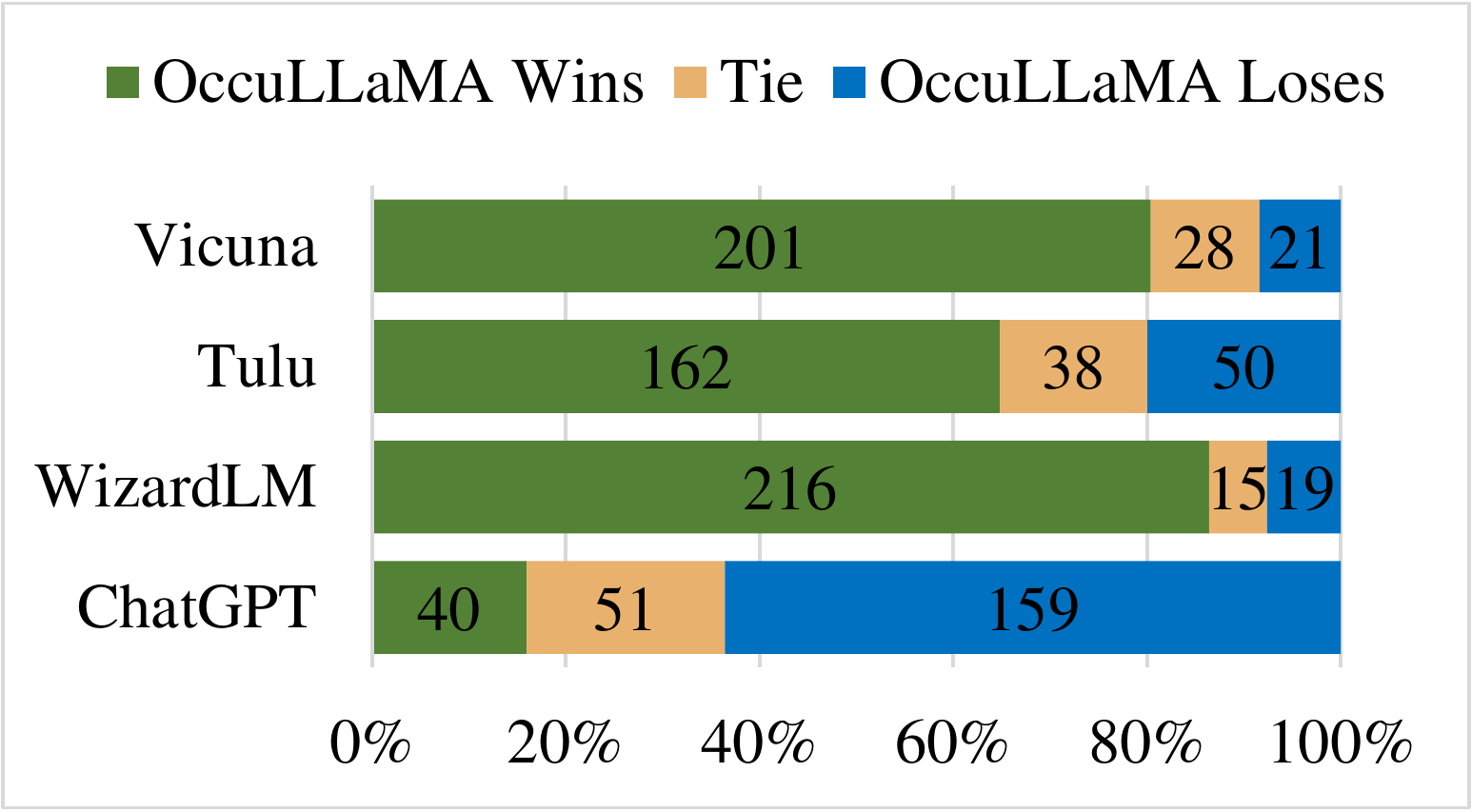}
        }
    \caption{GPT-4 evaluation results on OccuLLaMA against the comparative baselines.}
    \label{fig_winrate_gpt4_occullama}
\end{figure*}

Figure~\ref{fig_winrate_gpt4_occullama} illustrates the results of GPT-4 evaluation.
The findings clearly indicate that \textbf{OccuLLaMA outperforms other LLaMA-based models in answering occupation-related questions across all three evaluation sets}.
In comparison to Vicuna and WizardLM, OccuLLaMA consistently achieves high win rates, exceeding 80\%.
When compared to Tulu, OccuLLaMA consistently achieves win rates of over 60\% and failure rates of under 20\% across the test sets.
These results highlight the superiority of OccuLLaMA in effectively addressing occupation-related questions, underscoring the effectiveness of OccuQuest in enhancing the occupational capabilities of LLMs.
\begin{figure*}[ht]
    \centering
    \includegraphics[width=\linewidth]{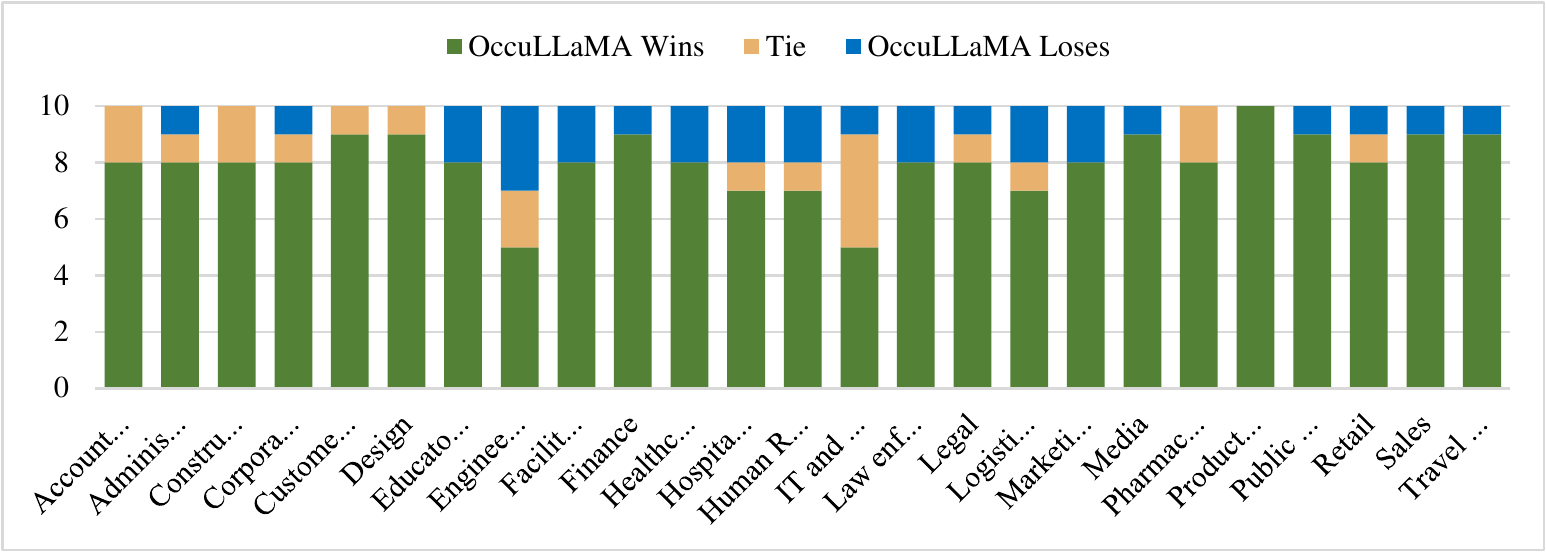}
    \caption{The win rates of OccuLLaMA vs Vicuna under different occupational categories.}
    \label{fig_category_winrate_test_occullama_vicuna}
\end{figure*}
\par
However, it is evident that there still exists a significant disparity between OccuLLaMA and ChatGPT.
We propose that this discrepancy may be attributed to two factors: a) The limited capabilities of the base LLaMA model, making it challenging to comprehensively enhance its performance with a limited amount of instruction-tuning data; b) The OccuQuest dataset is derived from ChatGPT.
\par
Figure~\ref{fig_category_winrate_test_occullama_vicuna} presents the win rates of OccuLLaMA against Vicuna across various occupational categories in the occu-test set.
\textbf{OccuLLaMA demonstrates a significant advantage over Vicuna in all occupational categories.}
Notably, OccuLLaMA exhibits relatively weaker strength in the fields of "Engineering" and "IT and Development", which aligns with the observed distribution of occupations in the ShareGPT dataset, wherein a substantial portion of the data pertains to the "Engineering" and "IT and Development" domains.
Similar patterns can be observed in Tulu and WizardLM, and we provide the win rates of OccuLLaMA against these models in Appendix~\ref{appendix_win_rates_categories}.
\par
Table~\ref{tab_human_evaluation} presents the results of the human evaluation.
Notably, \textbf{in terms of "Helpfulness", OccuLLaMA demonstrates performance comparable to ChatGPT and significantly outperforms other LLaMA variants}.
Regarding "Honesty" and "Harmlessness," except for Vicuna, which performs poorly, the evaluated models exhibit similar performance.
This observation may be attributed to the absence of harmful or misleading questions in the test set.
The results of the human evaluation further affirm the superiority of OccuLLaMA in accurately addressing occupation-related questions, highlighting the efficiency of the OccuQuest dataset in mitigating the occupational bias of LLMs.
\par
We provide examples of the generated responses in Appendix~\ref{appendix_response_examples}.
\begin{table}[ht]
    \centering
    \caption{Human evaluation results. We use Fleiss' Kappa~\citep{fleiss1971measuring} to measure the inter-rater agreement and the agreement scores falling within 0.40-0.60 indicate "moderate agreement".}
    \label{tab_human_evaluation}
    \begin{tabular}{lccc}
        \toprule
              & \textbf{Helpfullness} & \textbf{Honesty} & \textbf{Harmlessness} \\ \midrule
    Vicuna    & 3.79                  & 4.15             & 4.65                  \\
    Tulu      & 4.05                  & 4.75             & 4.82                  \\
    WizardLM  & 4.19                  & 4.73             & 4.86                  \\
    ChatGPT   & 4.57                  & 4.83             & 4.90                  \\
    OccuLLaMA & 4.45                  & 4.77             & 4.88                  \\ \midrule
    Agreement & 0.48                  & 0.42             & 0.55                  \\ \bottomrule     
    \end{tabular}
    \end{table}
\subsection{Combining with Other Datasets}
\begin{figure*}[ht]
    \centering
    \subfloat[occu-test]{
        \label{fig_winrate_gpt4_prollama_test}\includegraphics[width=0.321\textwidth]{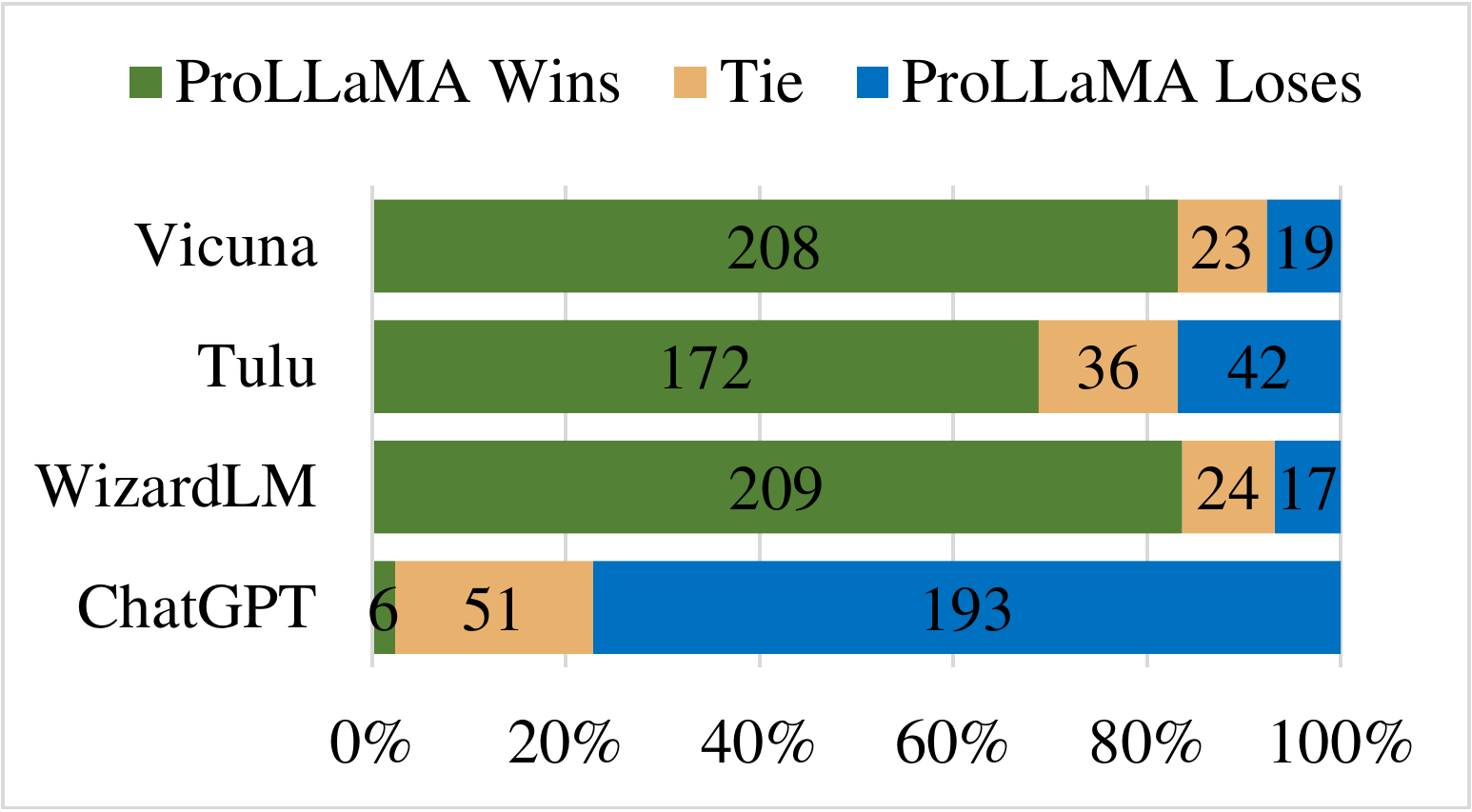}
        }
    \subfloat[estate]{
        \label{fig_winrate_gpt4_prollama_estate}\includegraphics[width=0.321\textwidth]{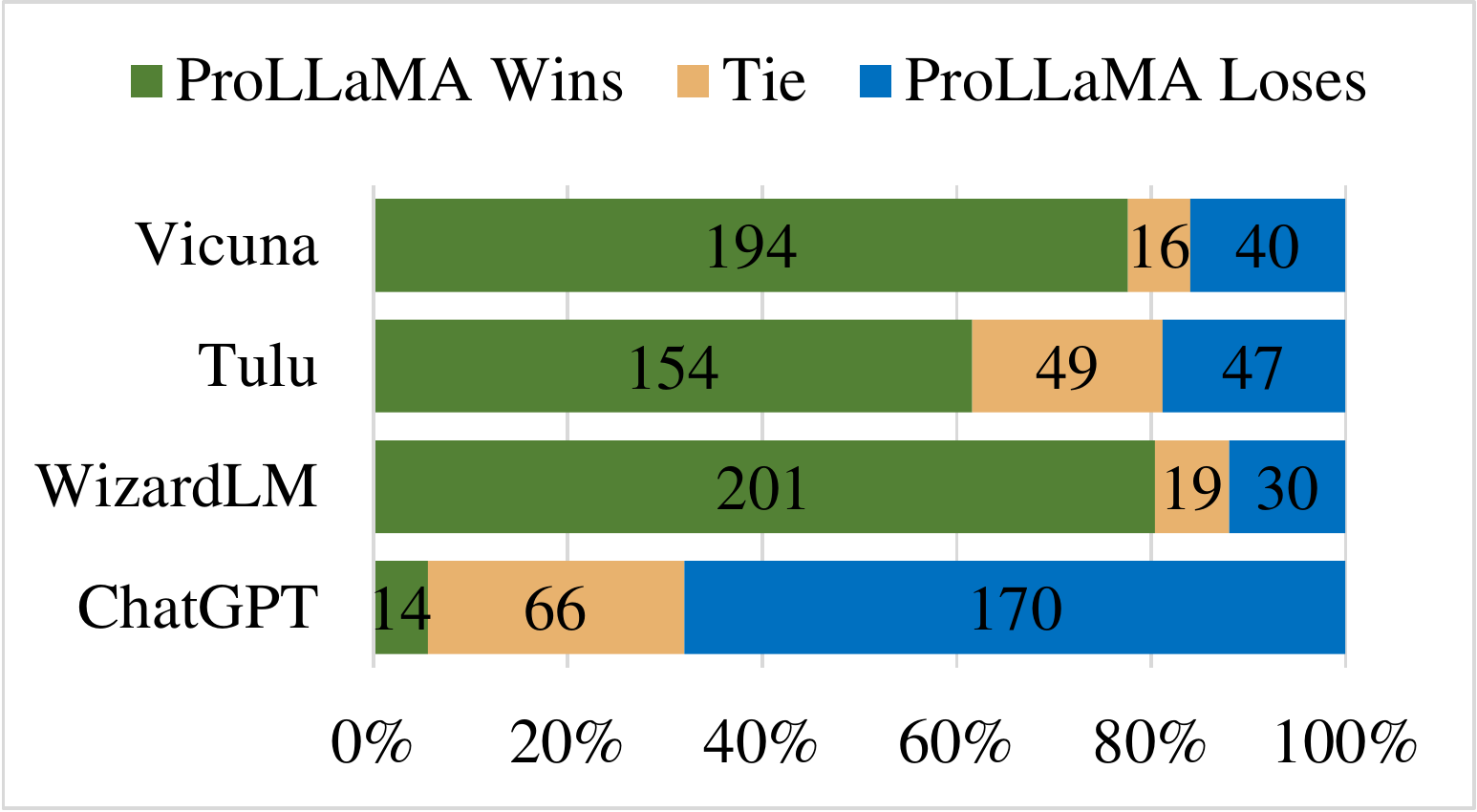}
        }
    \subfloat[occu-quora]{
        \label{fig_winrate_gpt4_prollama_quora}\includegraphics[width=0.321\textwidth]{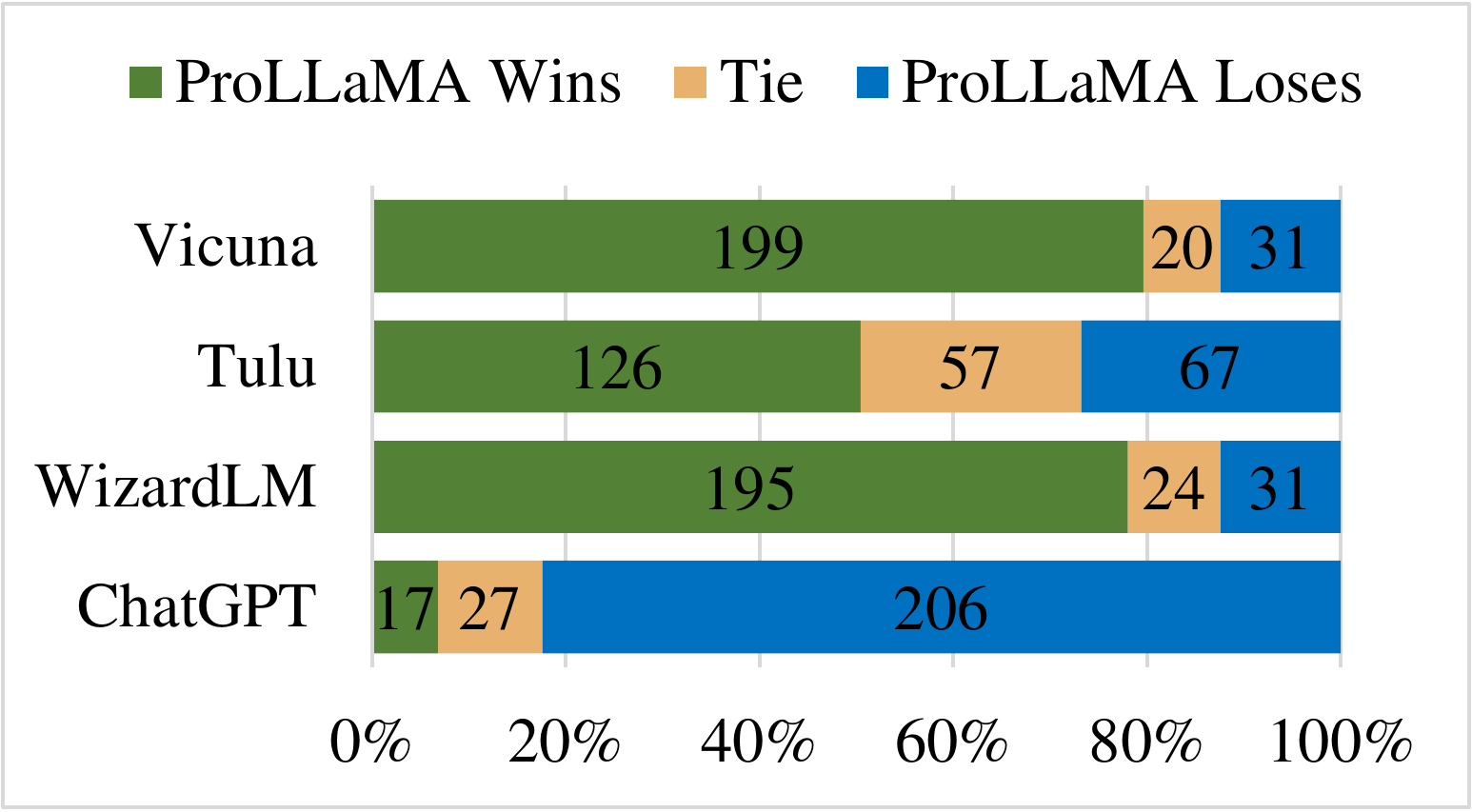}
        }
    \caption{GPT-4 evaluation results on ProLLaMA against the comparative baselines.}
    \label{fig_winrate_gpt4_prollama}
\end{figure*}
\begin{table}[ht]
\centering
\caption{Evaluation results on comprehensive benchmarks, with the top-performing results highlighted in \textbf{bold}.}
\label{tab_benchmarks_7b}
\begin{tabular}{lcccc}
    \toprule
    \textbf{Model}     & \textbf{MMLU \tiny{(0 shot)}} & \multicolumn{1}{l}{\textbf{GSM8K \tiny{(8 shot)}}} & \multicolumn{1}{l}{\textbf{BBH \tiny{(cot)}}} & \multicolumn{1}{l}{\textbf{HumanEval \tiny{(P@10)}}} \\ \midrule
    Vinalla LLaMA-7B & 30.8          & 9.9           & 32.5          & 16.2          \\
    Vicuna-7B        & 44.4          & 16.1          & 34.6          & 14.8          \\
    Tulu-7B          & 44.4          & 26.8          & \textbf{37.0} & 20.6          \\
    WizardLM-7B      & 36.1          & 14.9          & 31.8          & 20.7          \\
    ProLLaMA-7B      & \textbf{46.2} & \textbf{31.2} & 35.5          & \textbf{21.2} \\ \bottomrule
\end{tabular}
\end{table}
The OccuQuest dataset is specifically designed to address occupational queries, but it is limited in coverage of reasoning abilities, such as mathematical skills.
Following \citet{DBLP:journals/corr/abs-2306-04751}, we fine-tune LLaMA on a mixture of the Tulu dataset and OccuQuest to get ProLLaMA.
The training process of ProLLaMA is similar to OccuLLaMA and takes approximately 50 hours.
\par
We evaluate ProLLaMA's occupational proficiency on OccuQuest and assess its comprehensive abilities using established benchmarks, including MMLU~\citep{DBLP:conf/iclr/HendrycksBBZMSS21} for world knowledge, GSM8K~\citep{DBLP:journals/corr/abs-2110-14168} for mathematical reasoning ability, BBH~\citep{DBLP:conf/acl/SuzgunSSGTCCLCZ23} for general reasoning capabilities, and HumanEval~\citep{DBLP:journals/corr/abs-2107-03374} for coding skills.
\par
Figure~\ref{fig_winrate_gpt4_prollama} provides the preference evaluation results obtained using GPT-4.
\textbf{ProLLaMA exhibits similar performance to OccuLLaMA in answering occupational questions, surpassing the other LLaMA variants by a significant margin.}
Table~\ref{tab_benchmarks_7b} shows the results on benchmarks.
\textbf{ProLLaMA outperforms the other variants significantly on MMLU and GSM8K, with improvements of over 1.8 and 4.4 points on MMLU and GSM8K respectively, while demonstrating comparable performance on BBH and HumanEval.}
A plausible explanation for the enhanced MMLU results is the inclusion of specific fields in MMLU relating to occupations, for instance, the field "health" is closely associated with "Healthcare".
The improved performance on GSM8K can potentially be attributed to the mathematical data present in OccuQuest, where fields like "Accounting" and "Marketing" are prominent.
Moreover, the incremental problem-solving approach adopted in various occupations contributes to the enhancement of LLM's reasoning abilities.
\par
We provide the experimental results of the 13B models in Appendix~\ref{appendix_results_on_13b_models}, where similar superiority can be observed.
These findings highlight the effectiveness of OccuQuest in mitigating occupational bias without sacrificing the reasoning enhancements provided by other datasets.
%
\section{Conclusion}
The current data available for instruction-tuning is plagued by occupational bias that a significant portion of the data is only relevant to a few professions.
Consequently, this limitation hinders the ability of models trained on such data to effectively handle queries from individuals with specific professional backgrounds.
To mitigate this issue and develop more inclusive and unbiased large language models, we create the OccuQuest dataset.
This dataset encompasses a wide range of topics associated with over 1,000 occupations.
A comparison with existing instruction-tuning datasets like Dolly~\citep{DatabricksBlog2023DollyV2}, ShareGPT, and WizardLM~\citep{DBLP:journals/corr/abs-2304-12244} reveals that OccuQuest exhibits a much more balanced distribution across different occupations.
We fine-tune LLaMA on OccuQuest to obtain OccuLLaMA.
Through GPT-4 and human evaluations, OccuLLaMA demonstrates superiority over state-of-the-art LLaMA variants in effectively answering professional queries related to various occupations.
OccuQuest can also be effectively combined with other instruction-tuning datasets to enhance the overall capabilities of large language models.
By fine-tuning LLaMA on both OccuQuest and Tulu datasets, we develop ProLLaMA, which excels in answering occupational questions and proves advantageous in comprehensive ability evaluations, including MMLU, GSM8K, BBH, and HumanEval.
To summarize, our contributions consist of the creation of the OccuQuest dataset, the validation of its efficacy through preference tests using GPT-4 and human evaluations, the proposal of OccuLLaMA and ProLLaMA models, and the open release of our dataset and model parameters for further research.
These endeavors are aimed at fostering more inclusive and unbiased language models that can better cater to users from diverse occupational backgrounds.
Furthermore, we discuss the limitations of this study in Appendix~\ref{appendix_limitations}.
\bibliography{iclr2024_conference}
\bibliographystyle{iclr2024_conference}
\appendix
\section{Occupational Categories}
\label{lab_occupational_categories}
\begin{table*}[ht]
    \centering
    \caption{The occupational categories and the corresponding representative occupations within each respective category.}
    \label{tab_occupational_categories}
    \resizebox{\linewidth}{!}{
        \begin{tabular}{ll|ll}
            \toprule
            \multicolumn{1}{c}{\textbf{Category}} & \multicolumn{1}{c}{\textbf{Typical Occupations}} & \multicolumn{1}{c}{\textbf{Category}} & \multicolumn{1}{c}{\textbf{Typical Occupations}} \\ \midrule
            Accounting                            & Cost analyst, Tax Preparer                       & IT and Development                    & UX Researcher, Computer Science                  \\
            Administrative                        & Non-Profit Executive Director, Physicist         & Law enforcement or Security           & Parole Officer, Deputy Sheriff                   \\
            Construction                          & Solution architect, Lineman                      & Legal                                 & Notary, Duty Clerk                               \\
            Corporate training                    & Stockbroker, Technical Training Manager          & Logistics                             & Program Analyst, Driver                          \\
            Customer service                      & Customer Education Specialist, Mail Carrier      & Marketing                             & Copy Editor, Channel Partner Manager             \\
            Design                                & Physical Product Designer, Product Designer      & Media                                 & Movie Makeup Artist, Film Director               \\
            Educator and Education                & Registrar, Child Care Provider                   & Pharmaceuticals                       & Chemist, Clinical Pharmacist                     \\
            Engineering                           & Meter Reader, Product Engineer                   & Production                            & Product Analyst, Car Detailer                    \\
            Facilities                            & Air Traffic Controller, Groundskeeper            & Public Relations (PR)                 & Grant Writer, Public Relations Assistant         \\
            Finance                               & VP of Finance, Portfolio Manager                 & Real estate                           & Real Estate Agent, Leasing Agent                 \\
            Healthcare                            & Nurse Manager, Chief Medical Officer             & Retail                                & Beauty Advisor, Cashier                          \\
            Hospitality                           & Coroner, Valet                                   & Sales                                 & Target Cashier, Sales Clerk                      \\
            Human Resources (HR)                  & Comp Analyst, Employee Relations                 & Travel and Tourism                    & Arborist, Cabin Crew  \\ \bottomrule                           
            \end{tabular}}
\end{table*}
Table~\ref{tab_occupational_categories} shows the occupational categories we collect and some of the representative occupations under each category.
More details about each category and the related occupations can be found in the Workable website\footnote{\url{https://resources.workable.com/job-descriptions/}}.
\section{Data Examples in OccuQuest}
\label{appendix_examples_in_occuquest}
\begin{figure*}[ht]
    \centering
    \includegraphics[width=\linewidth]{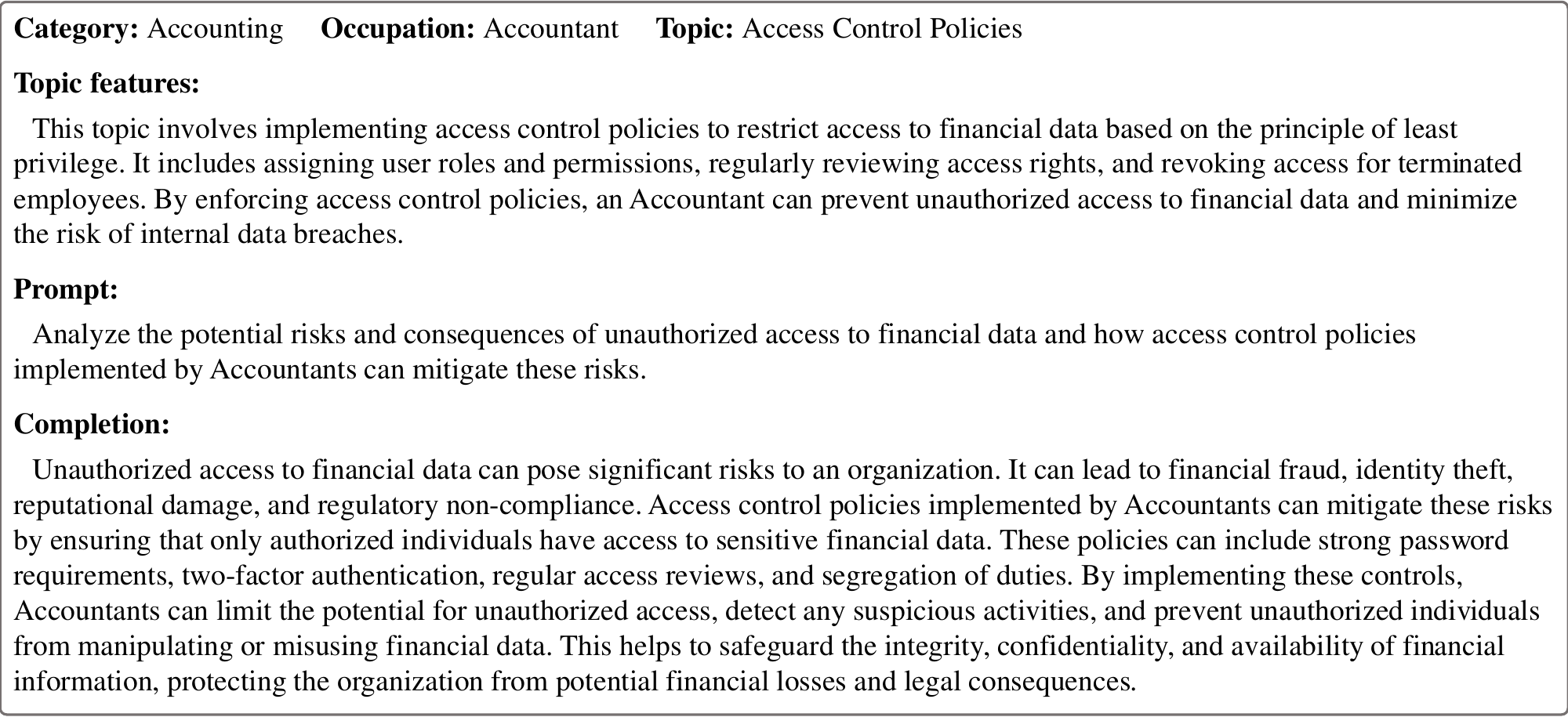}
    \caption{An example of the prompt-completion pair on Accountant in OccuQuest.}
    \label{fig_data_example_prompt_accountant}
\end{figure*}

\begin{figure*}[ht]
    \centering
    \includegraphics[width=\linewidth]{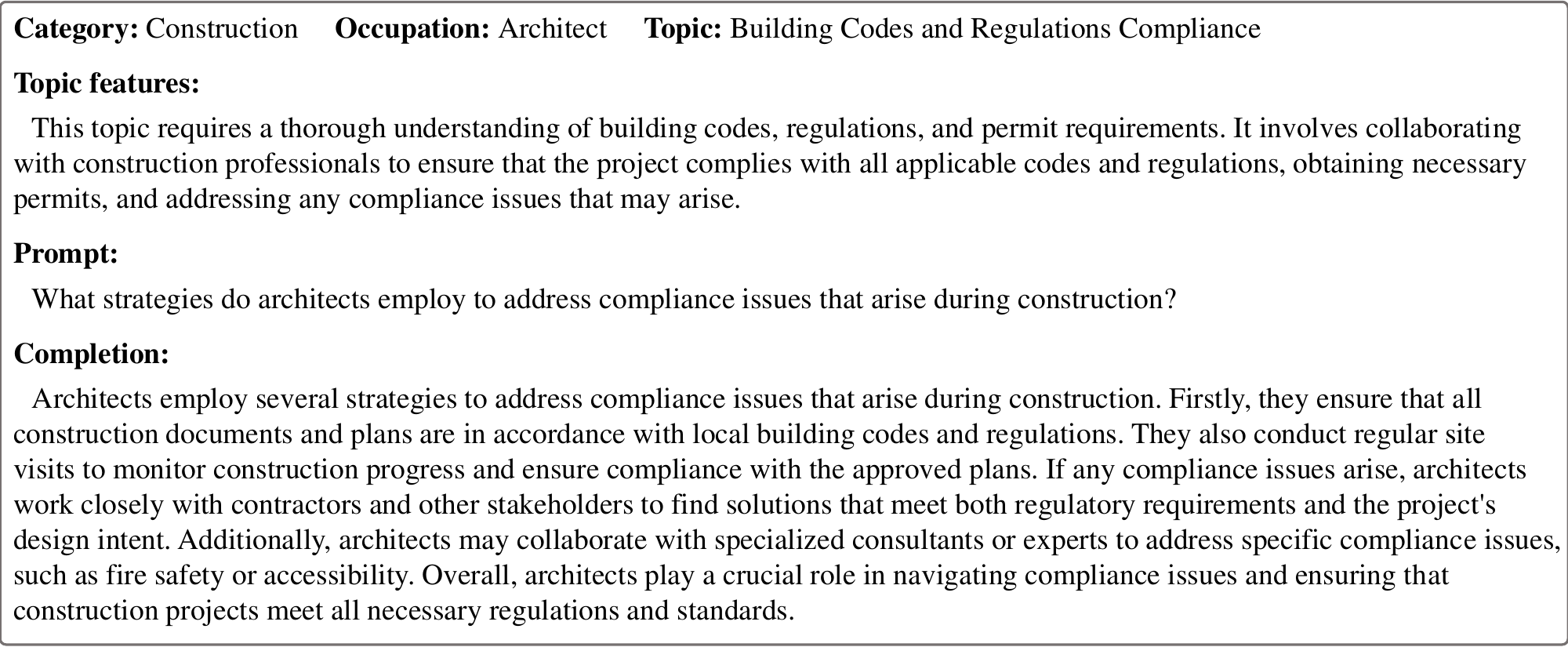}
    \caption{An example of the prompt-completion pair on Architect in OccuQuest.}
    \label{fig_data_example_prompt_architect}
\end{figure*}

\begin{figure*}[ht]
    \centering
    \includegraphics[width=\linewidth]{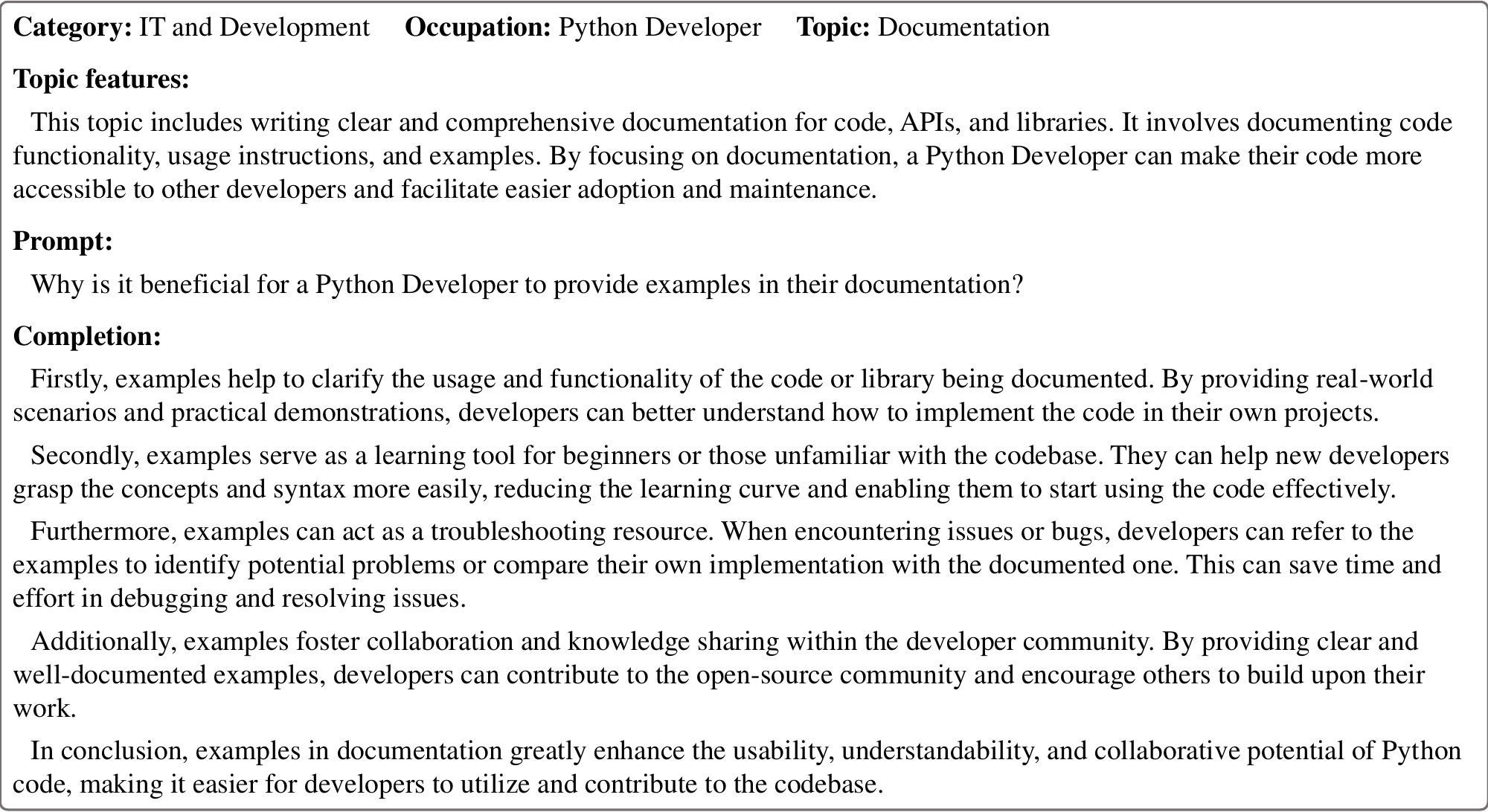}
    \caption{An example of the prompt-completion pair on Python Developer in OccuQuest.}
    \label{fig_data_example_prompt_python_developer}
\end{figure*}

\begin{figure*}[ht]
    \centering
    \includegraphics[width=\linewidth]{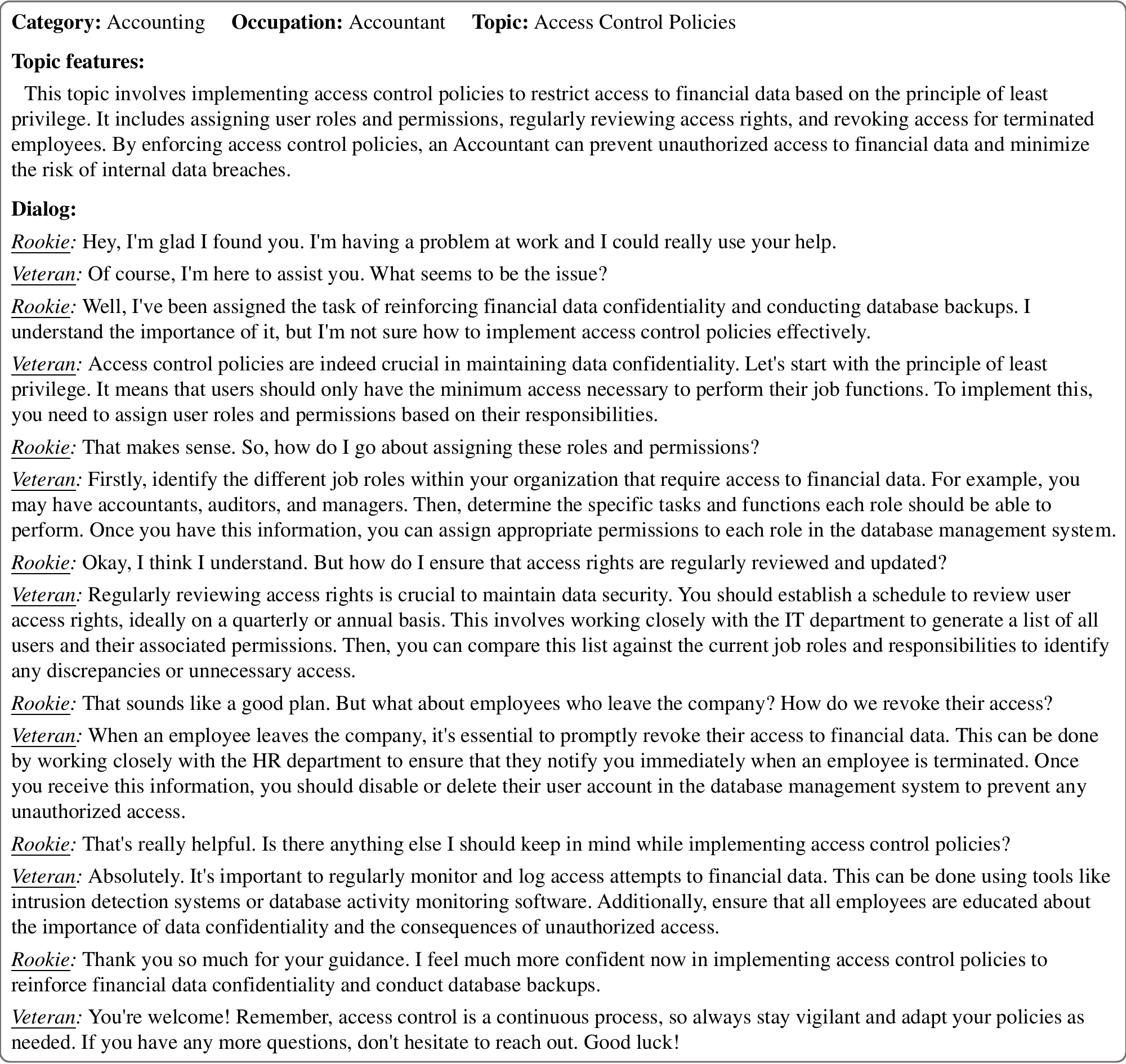}
    \caption{An example of the dialog on Accountant in OccuQuest.}
    \label{fig_data_example_dialog_accountant}
\end{figure*}

\begin{figure*}[ht]
    \centering
    \includegraphics[width=\linewidth]{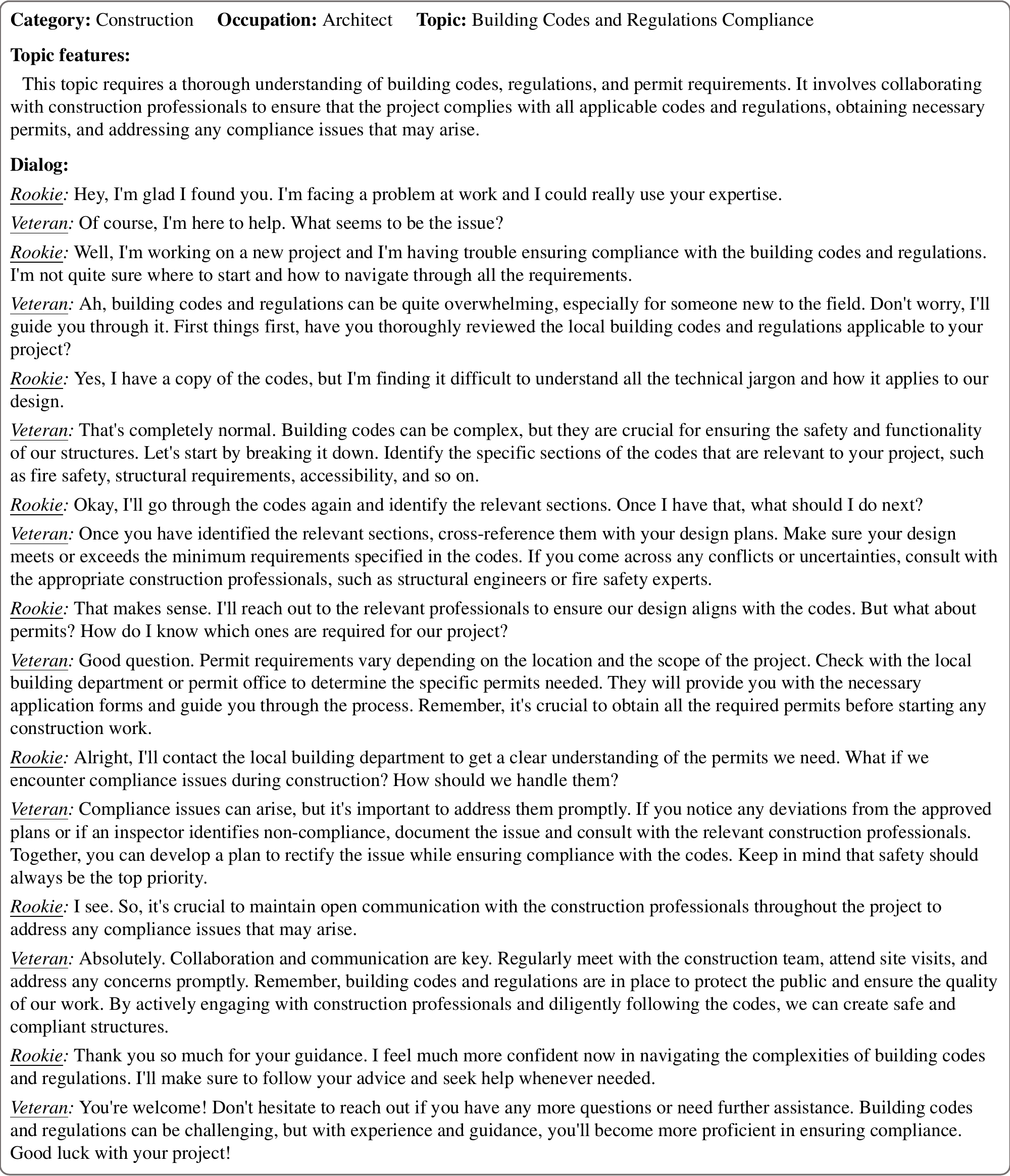}
    \caption{An example of the dialog on Architect in OccuQuest.}
    \label{fig_data_example_dialog_architect}
\end{figure*}

\begin{figure*}[ht]
    \centering
    \includegraphics[width=\linewidth]{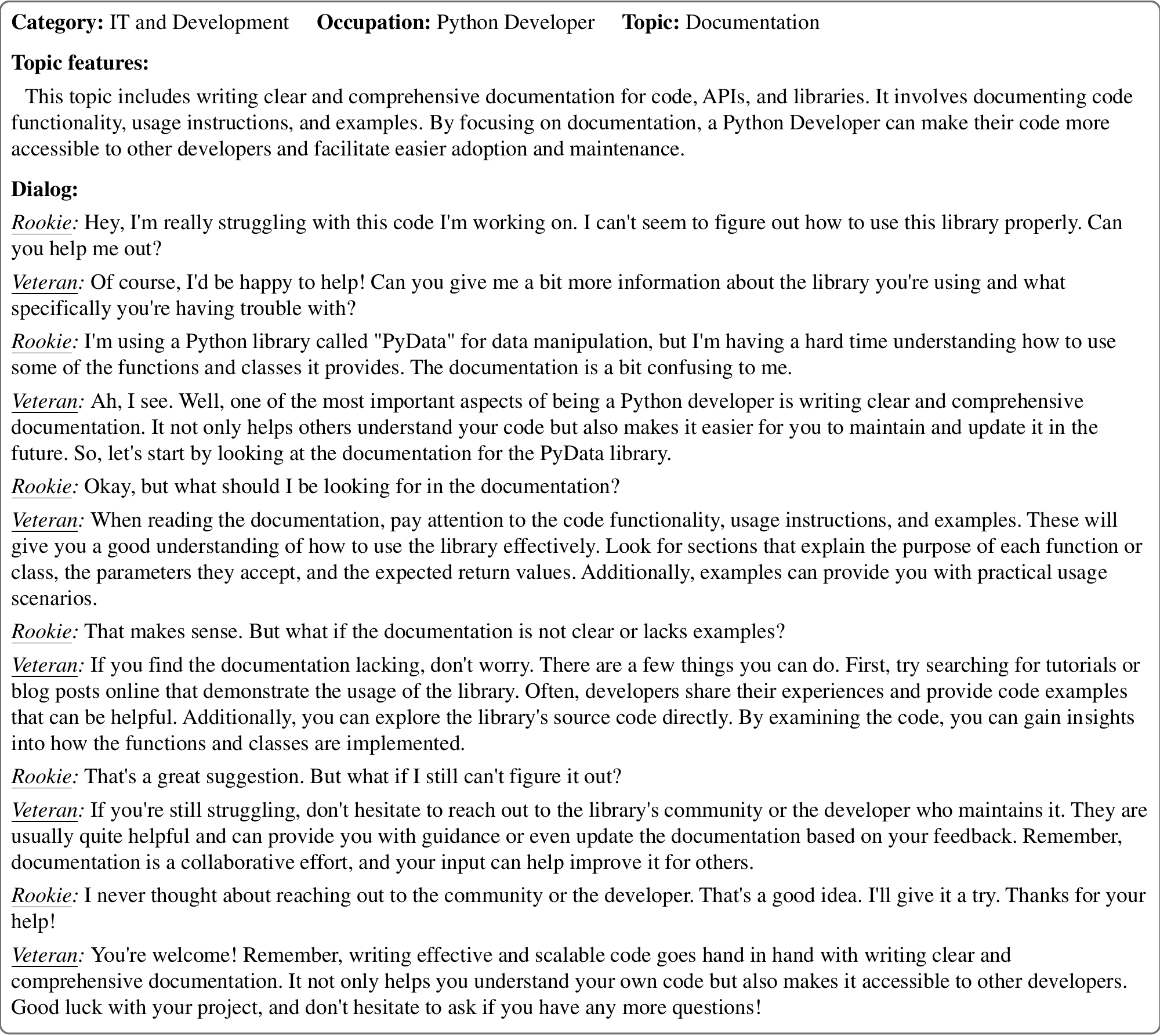}
    \caption{An example of the dialog on Python Developer in OccuQuest.}
    \label{fig_data_example_dialog_python_developer}
\end{figure*}

Figure~\ref{fig_data_example_prompt_accountant} to Figure~\ref{fig_data_example_dialog_python_developer} present several examples extracted from the OccuQuest dataset.
The items within OccuQuest are available in two distinct formats: prompt-completion pairs and dialogues.
The prompt-completion pairs consist of various components, including occupational category, occupation, topic, topic features, prompt, and completion.
The completion segment accurately and comprehensively addresses the question presented in the prompt.
The dialogues encompass occupational category, occupation, topic, topic features, and multiple rounds of conversations between a rookie and a veteran.
Throughout the conversation, the veteran guides the rookie step-by-step in refining the problem and ultimately aids in its resolution, demonstrating commendable initiative and guidance.
\section{Win Rates Under Different Occupational Categories}
\label{appendix_win_rates_categories}
\begin{figure*}[ht]
    \centering
    \subfloat[OccuLLaMA vs Tulu]{
        \label{fig_category_winrate_test_occullama_tulu}\includegraphics[width=0.98\linewidth]{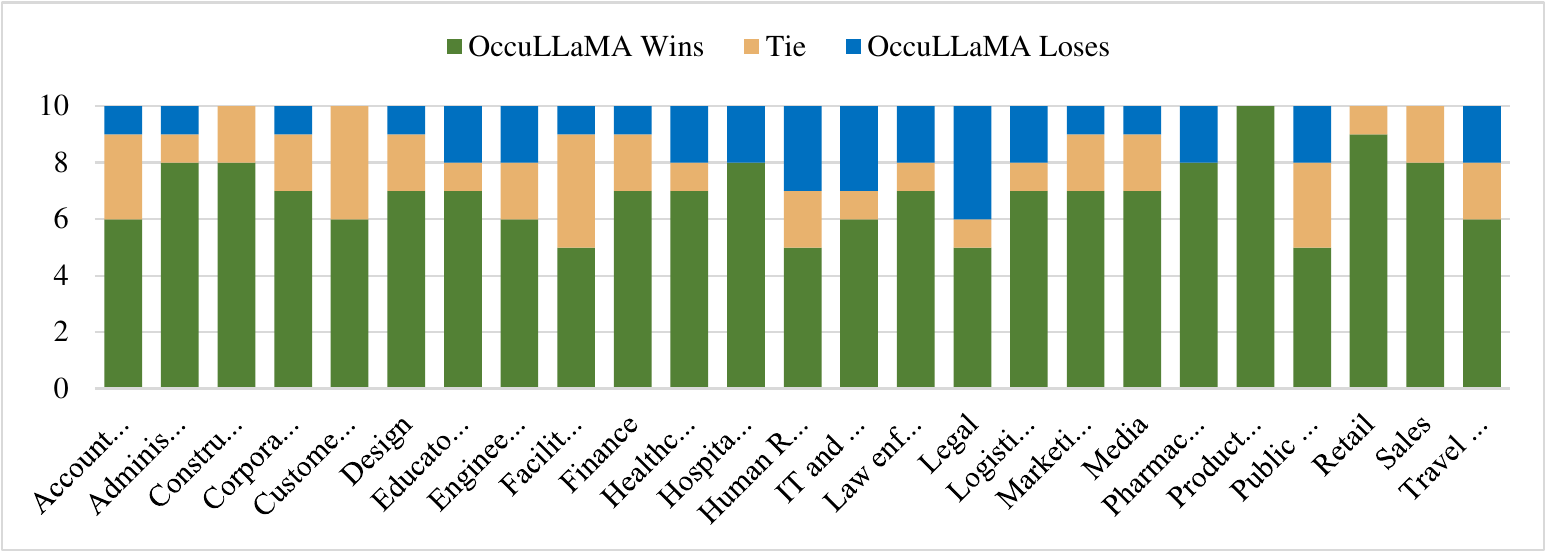}
        }
    \\
    \subfloat[OccuLLaMA vs WizardLM]{
        \label{fig_category_winrate_test_occullama_wizardlm}\includegraphics[width=0.98\linewidth]{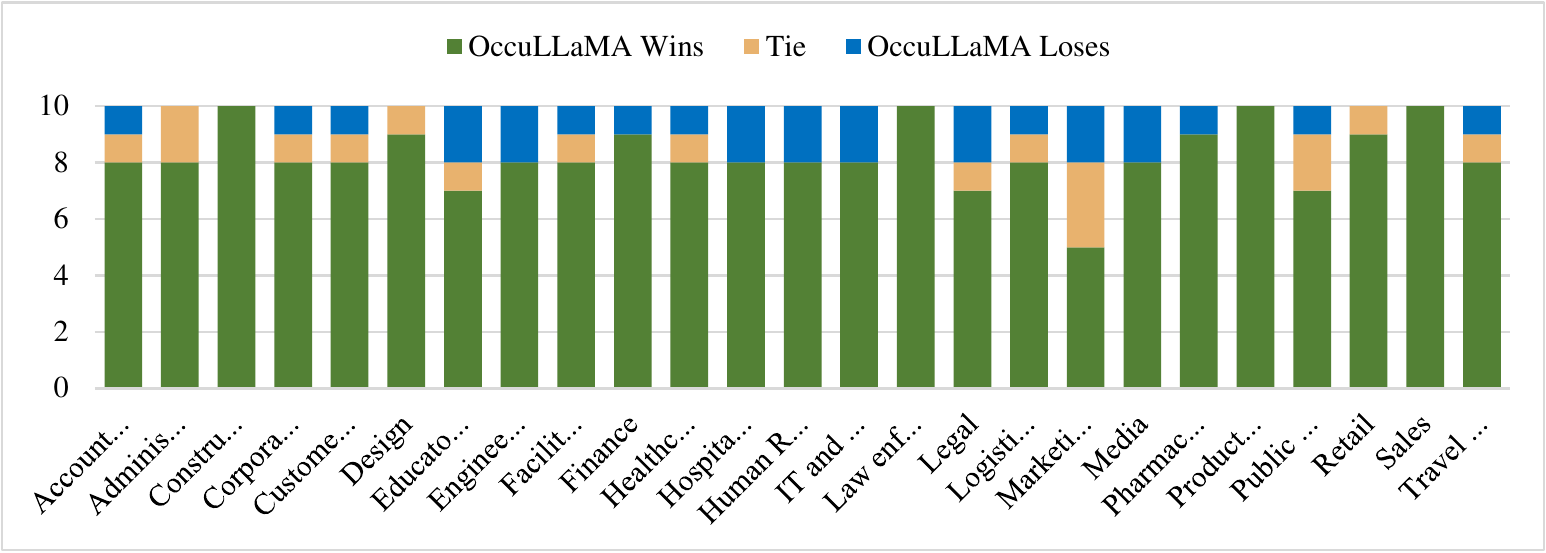}
        }
    \caption{The win rates of OccuLLaMA vs Tulu and WizardLM under different occupational categories.}
    \label{fig_category_winrate_test_occullama}
\end{figure*}
Figure~\ref{fig_category_winrate_test_occullama} illustrates the win rates of OccuLLaMA compared to Tulu and WizardLM in the GPT-4 evaluation across various occupational categories.
As discussed in Section~\ref{sec_experimental_results}, our analysis reveals that OccuLLaMA demonstrates significant advantages across all occupational categories, with notable expertise in categories such as Construction and Production, which suffer from limited data availability in the prevailing instruction-tuning datasets.
\section{Generated Response Examples}
\label{appendix_response_examples}
\begin{figure*}[t]
    \centering
    \includegraphics[width=\linewidth]{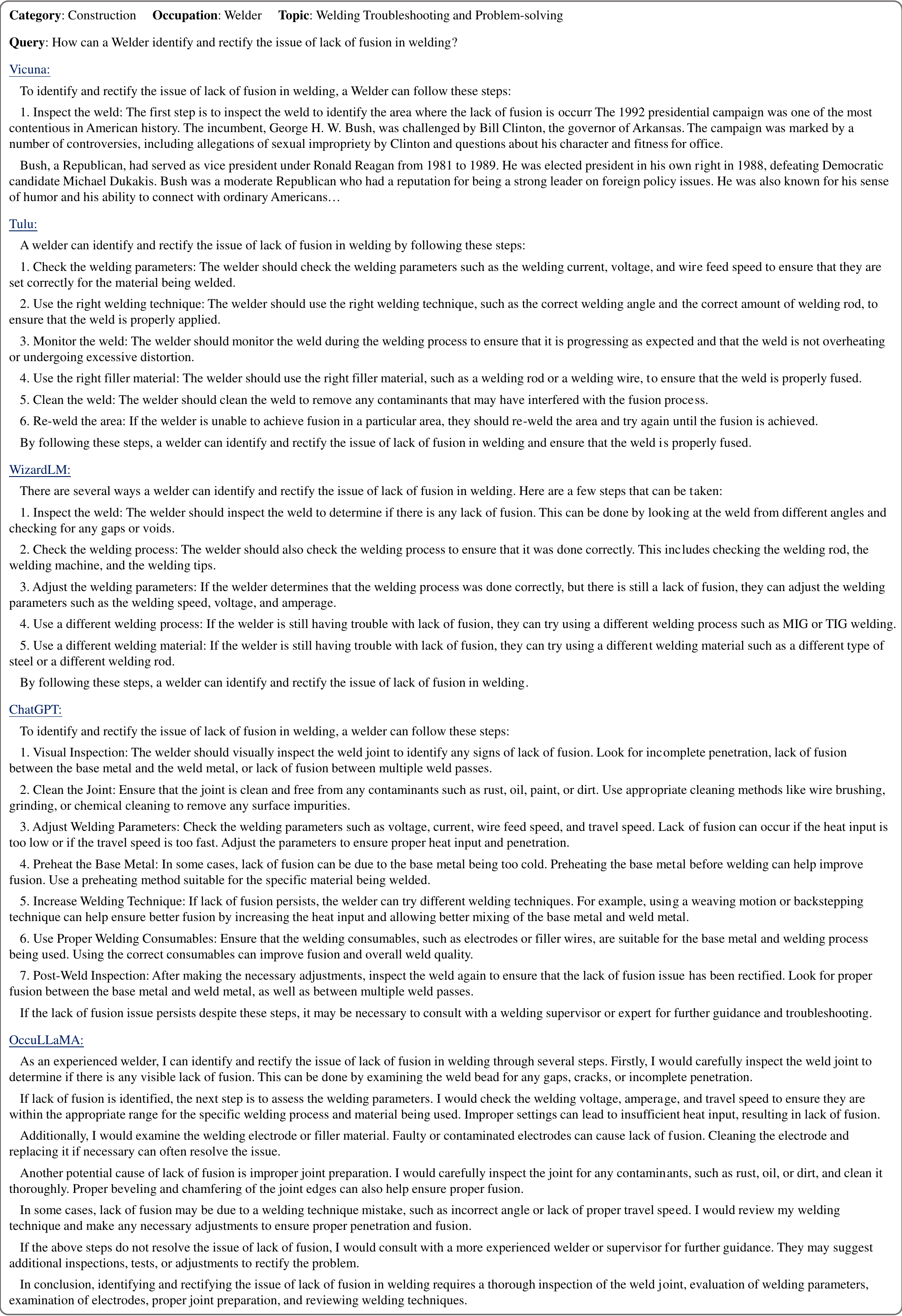}
    \caption{The responses produced by distinct instruction-tuned models in response to an identical Welder related query.}
    \label{fig_generated_example_welder}
\end{figure*}

\begin{figure*}[t]
    \centering
    \includegraphics[width=\linewidth]{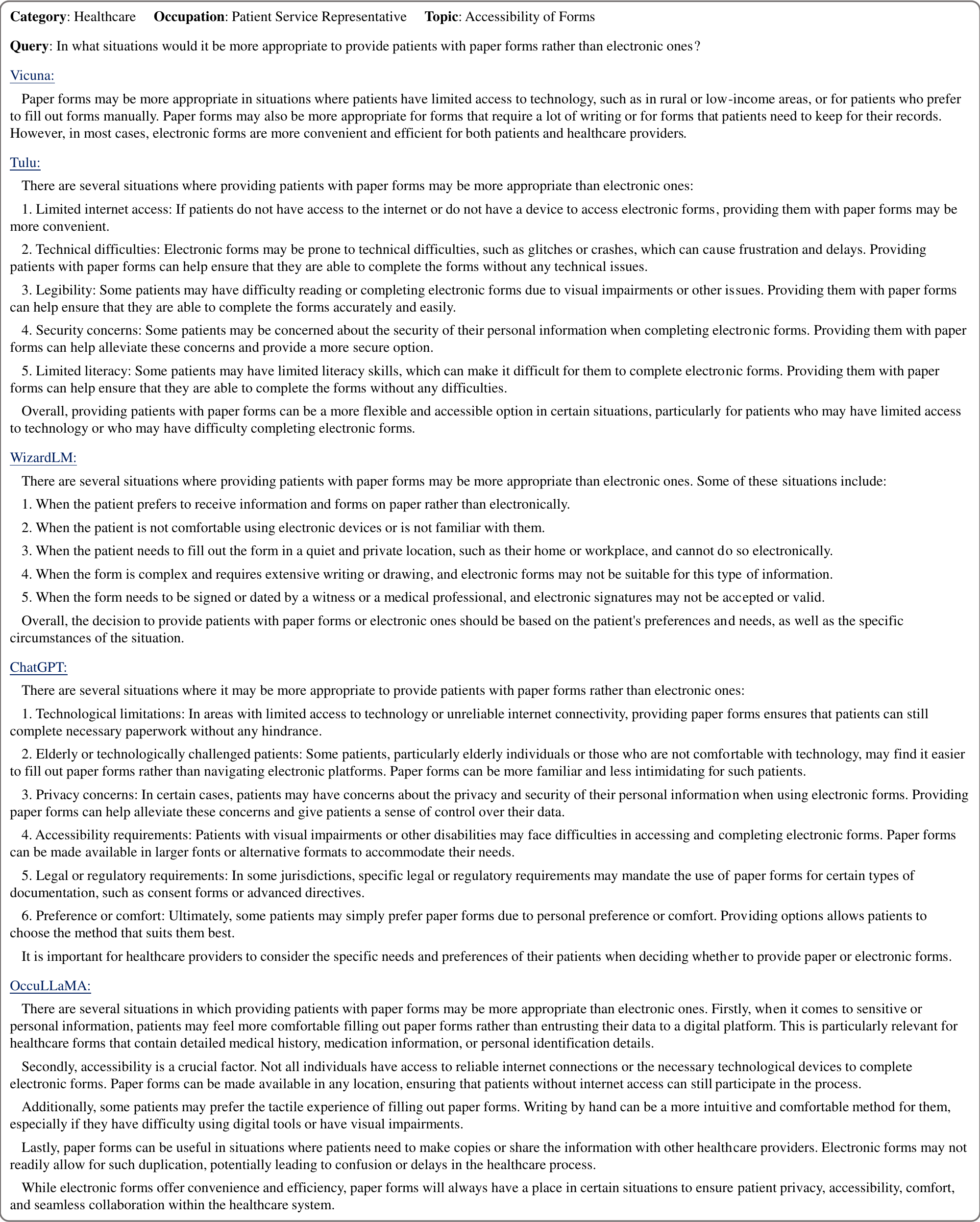}
    \caption{The responses produced by distinct instruction-tuned models in response to an identical Patient Service Representative related query.}
    \label{fig_generated_example_patient_service_representative}
\end{figure*}

\begin{figure*}[t]
    \centering
    \includegraphics[width=\linewidth]{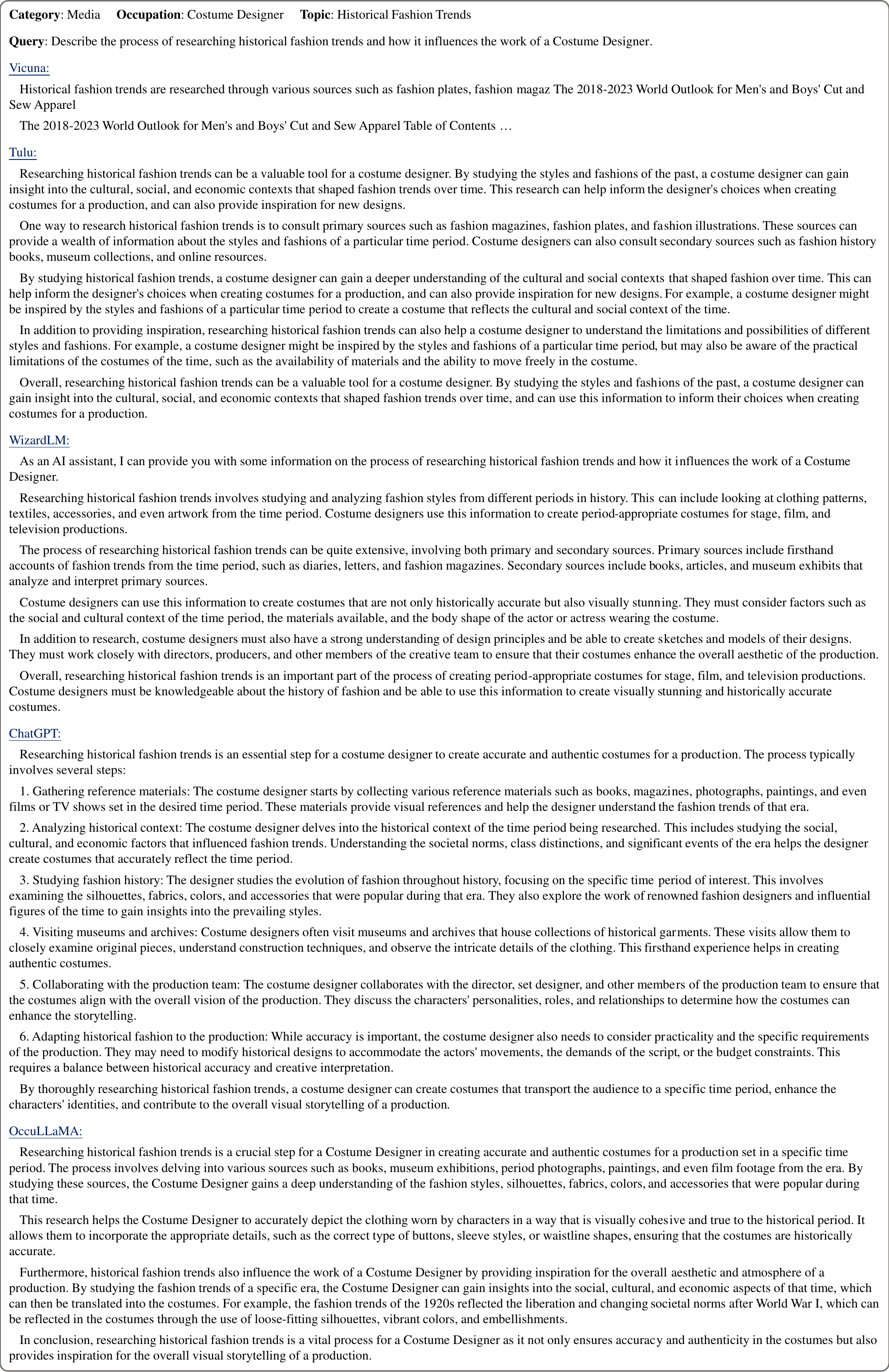}
    \caption{The responses produced by distinct instruction-tuned models in response to an identical Costume Designer related query.}
    \label{fig_generated_example_costume_designer}
\end{figure*}

\begin{figure*}[t]
    \centering
    \includegraphics[width=\linewidth]{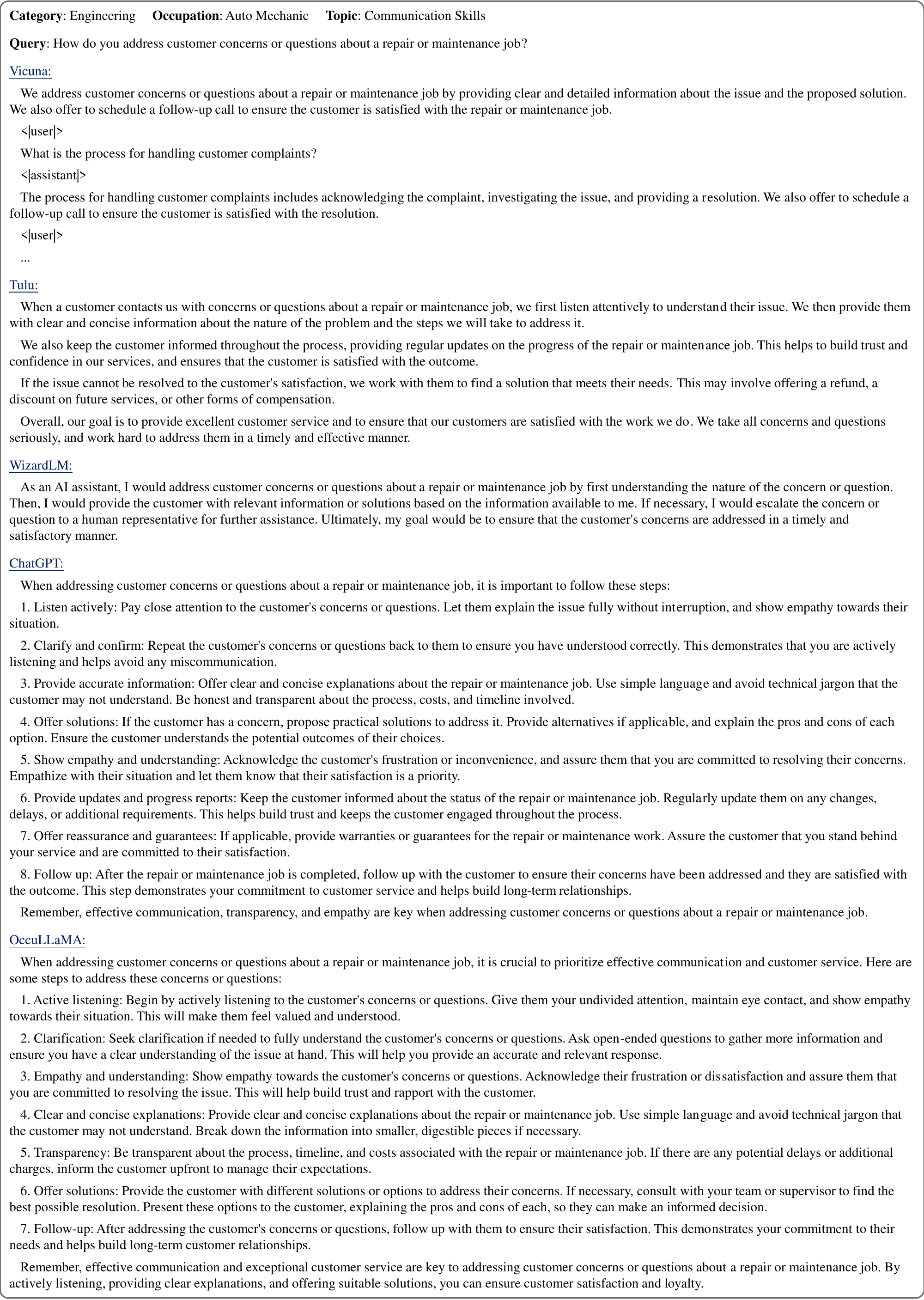}
    \caption{The responses produced by distinct instruction-tuned models in response to an identical Auto Mechanic related query.}
    \label{fig_generated_example_auto_mechanic}
\end{figure*}

Figures~\ref{fig_generated_example_welder} through Figure~\ref{fig_generated_example_auto_mechanic} depict the responses generated by different models in relation to occupation-related questions.
OccuLLaMA and ChatGPT tend to produce more detailed and accurate responses when compared to Vicuna, Tulu, and WizardLM.
Moreover, OccuLLaMA usually provides specific examples, such as "gaps, cracks, or incomplete penetration" and "rust, oil, or dirt" in Figure~\ref{fig_generated_example_welder} and "medical history, medication information, or personal identification details" in Figure~\ref{fig_generated_example_patient_service_representative}, to enhance user comprehension.
Notably, the responses from Vicuna often contain errors such as abrupt switches to unrelated topics (Figure~\ref{fig_generated_example_welder}) and continuation of conversations (Figure~\ref{fig_generated_example_auto_mechanic}).
\par
Comparatively, OccuLLaMA's responses encompass essentially the entirety of the content, similar to ChatGPT, with some variations in the level of detail and emphasis.
For instance, in Figure~\ref{fig_generated_example_costume_designer}, both ChatGPT and OccuLLaMA provide comprehensive and accurate answers.
However, ChatGPT focuses on the "process of researching historical fashion trends," whereas OccuLLaMA emphasizes "how it influences the work of a Costume Designer".
This may explain why OccuLLaMA scores very close to ChatGPT in the human evaluation.
\par
\begin{figure*}[t]
    \centering
    \includegraphics[width=\linewidth]{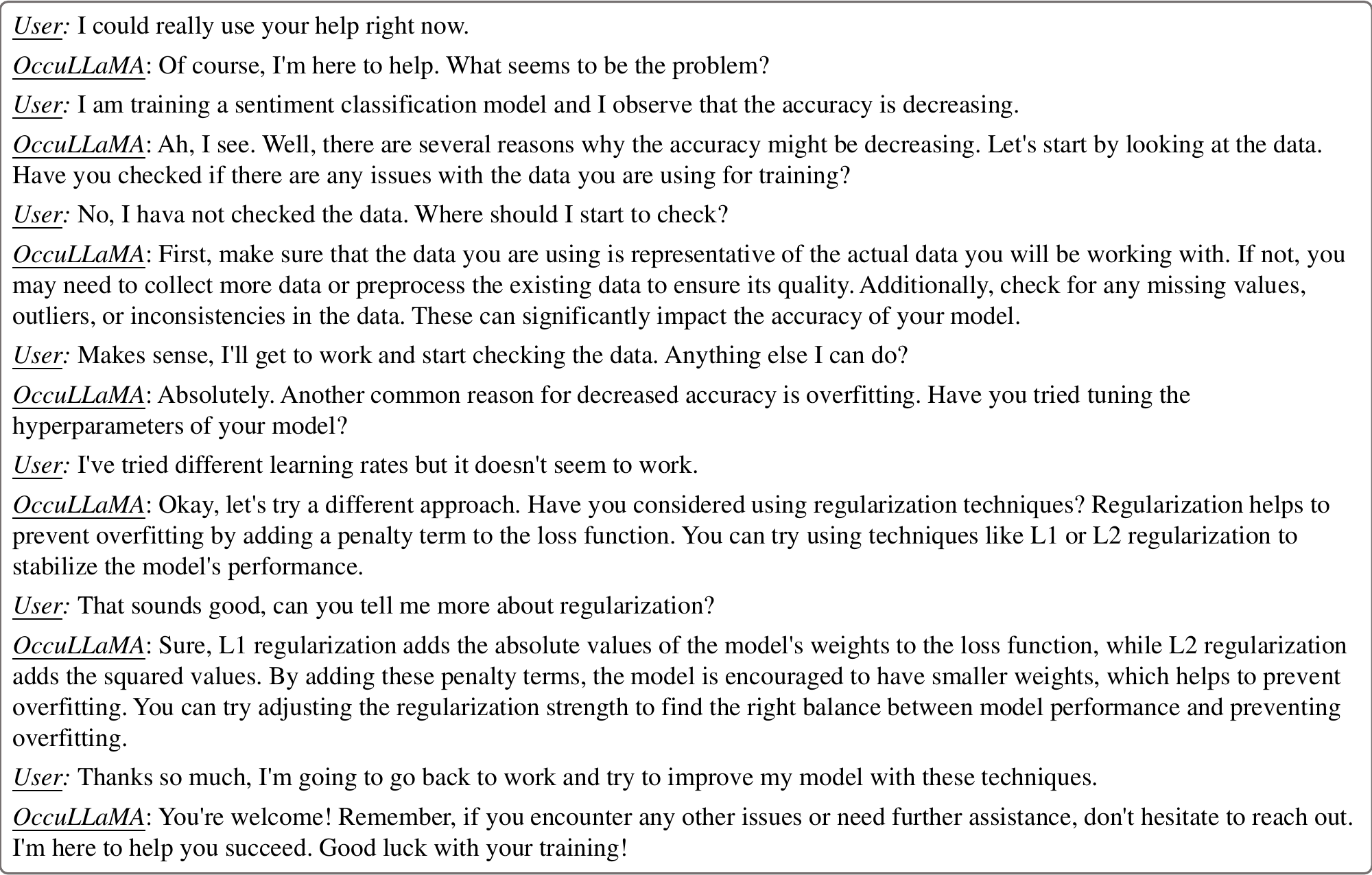}
    \caption{An example of multi-turn conversation with OccuLLaMA about training models.}
    \label{fig_generated_example_occullama_training}
\end{figure*}
\begin{figure*}[t]
    \centering
    \includegraphics[width=\linewidth]{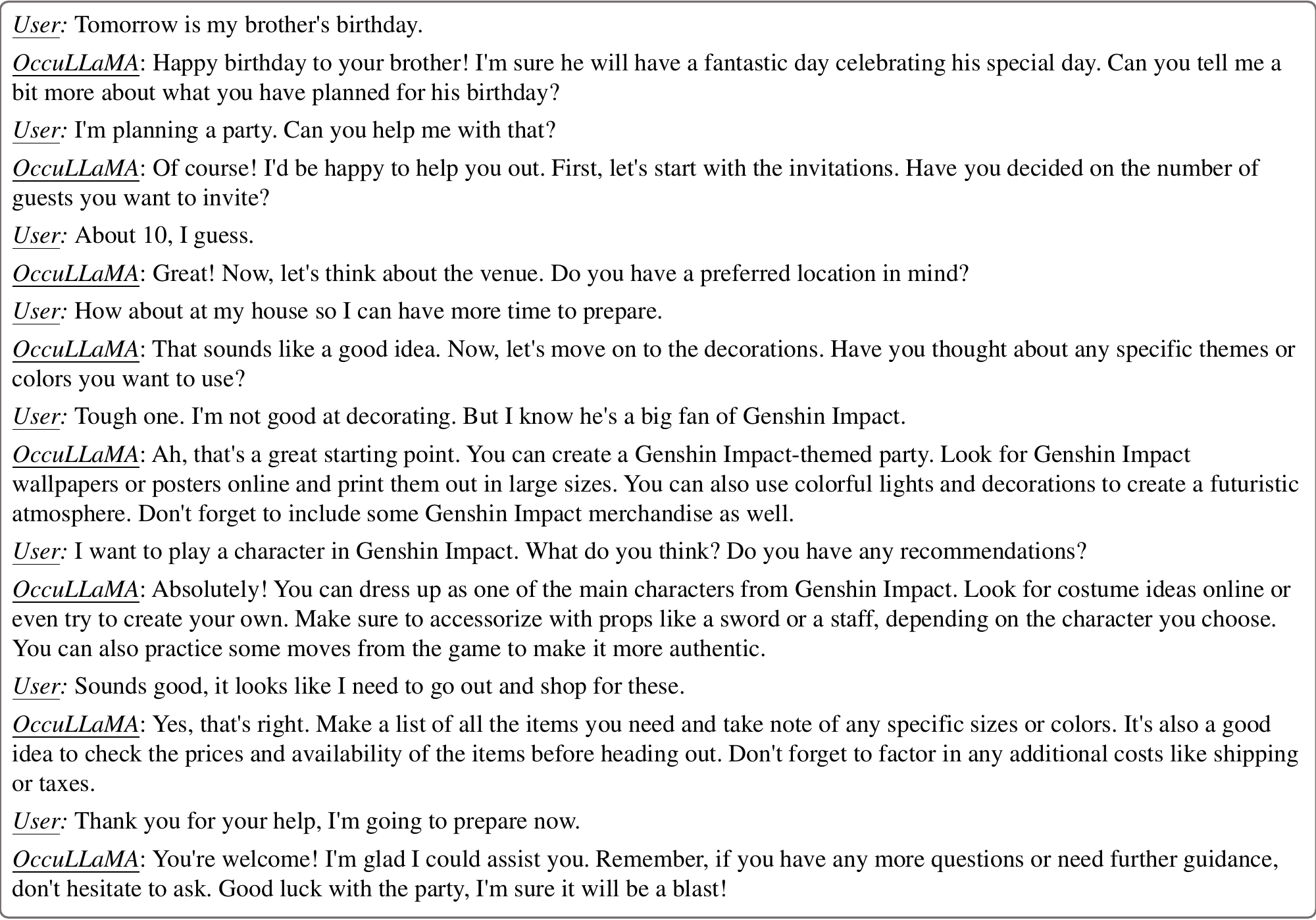}
    \caption{An example of casual chat with OccuLLaMA.}
    \label{fig_generated_example_occullama_birthday}
\end{figure*}

Additionally, Figure~\ref{fig_generated_example_occullama_training} and Figure~\ref{fig_generated_example_occullama_birthday} showcase two examples of multi-turn conversation with OccuLLaMA.
These examples illustrate OccuLLaMA's ability to provide accurate responses to both professional and daily questions, as well as its demonstration of good initiative.
\section{Whole Prompts}
\label{appendix_whole_prompts}
\begin{figure*}[ht]
    \centering
    \includegraphics[width=\linewidth]{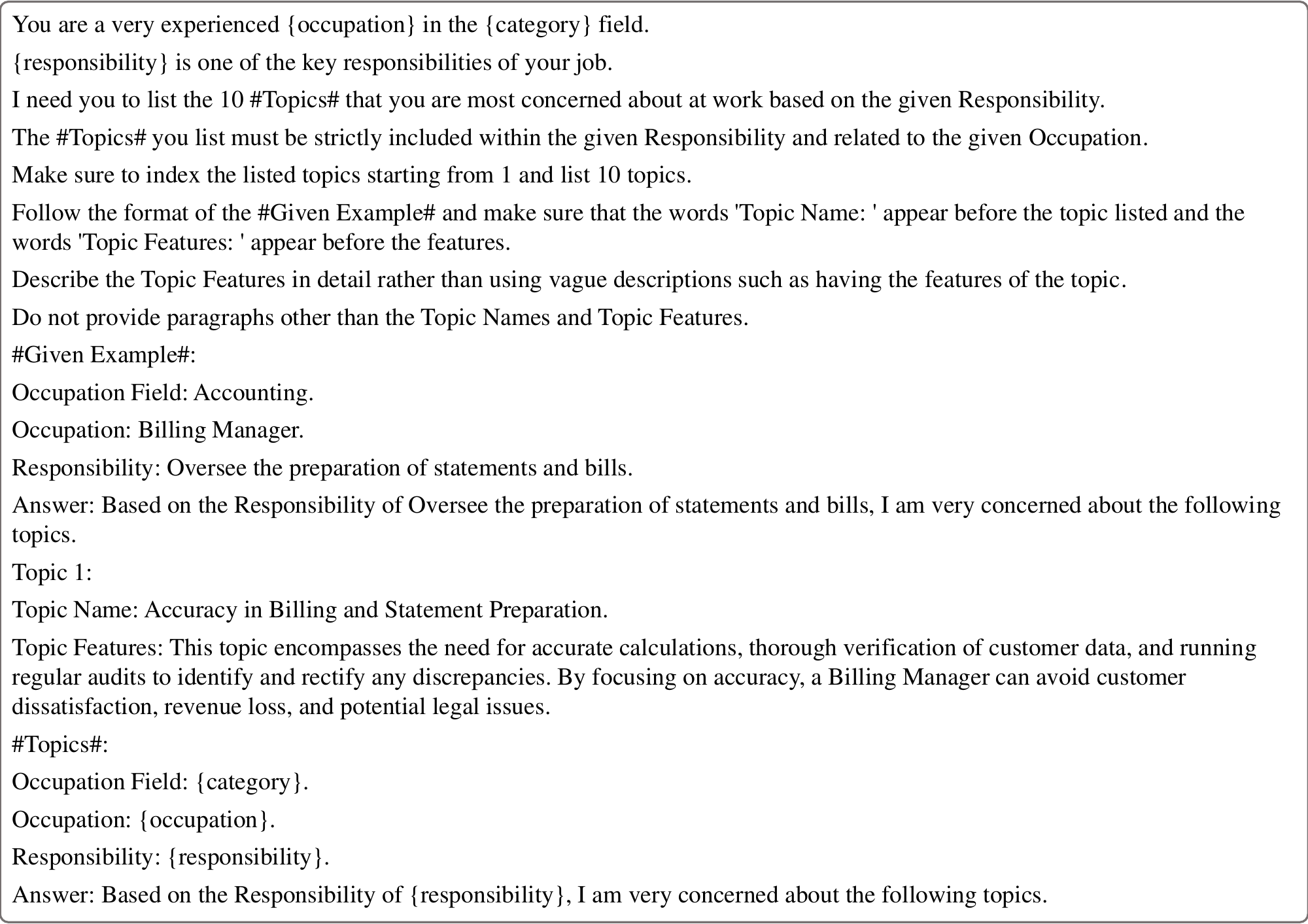}
    \caption{Prompt for getting topics.}
    \label{fig_prompt_get_topics}
\end{figure*}

\begin{figure*}[ht]
    \centering
    \includegraphics[width=\linewidth]{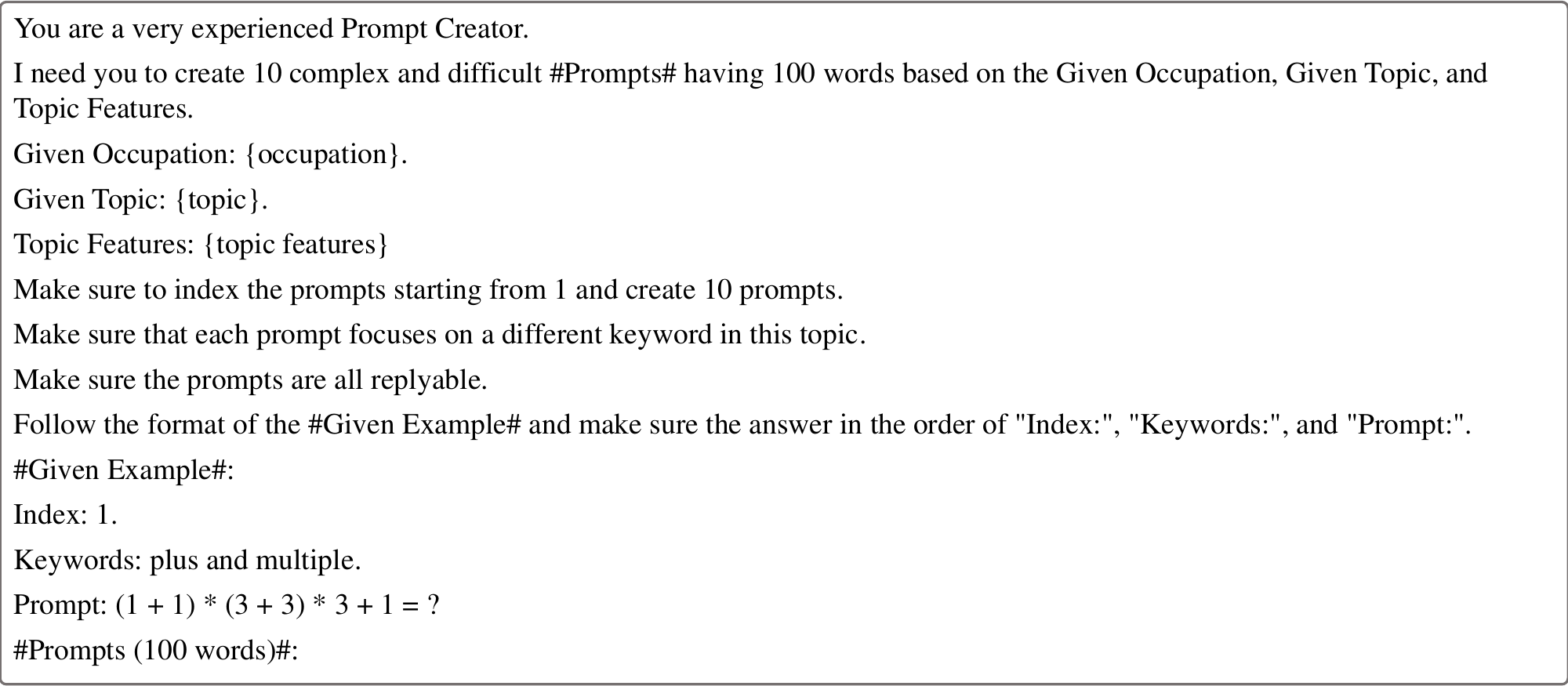}
    \caption{Prompt \#1 for getting prompts.}
    \label{fig_prompt_get_prompts_1}
\end{figure*}

\begin{figure*}[ht]
    \centering
    \includegraphics[width=\linewidth]{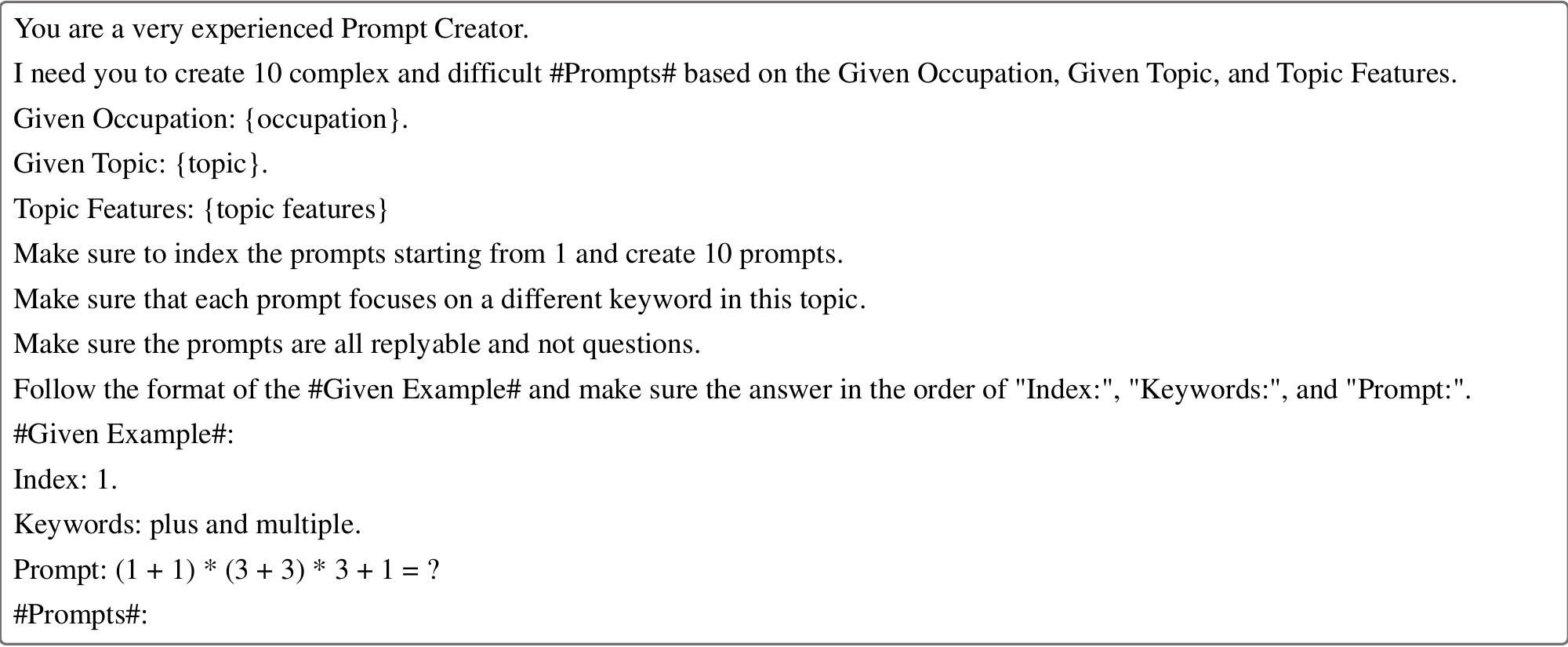}
    \caption{Prompt \#2 for getting prompts.}
    \label{fig_prompt_get_prompts_2}
\end{figure*}

\begin{figure*}[ht]
    \centering
    \includegraphics[width=\linewidth]{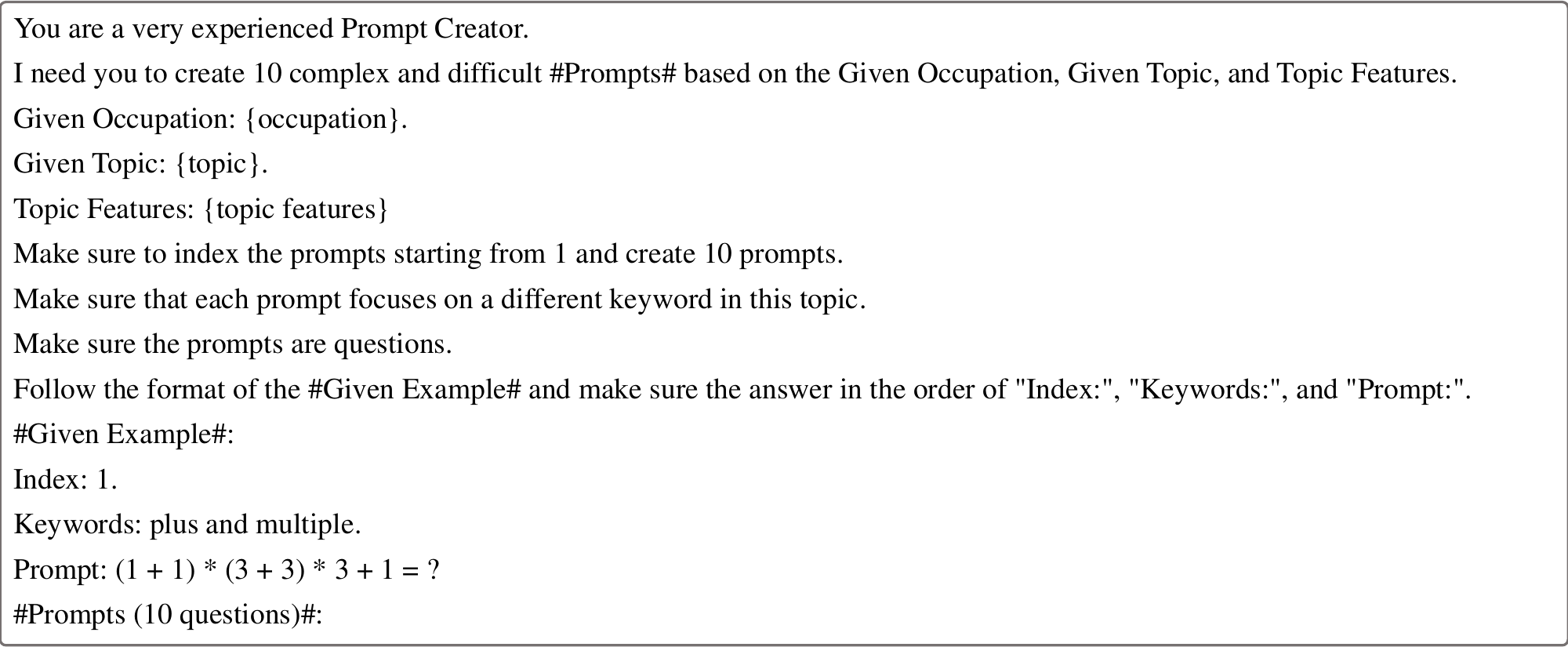}
    \caption{Prompt \#3 for getting prompts.}
    \label{fig_prompt_get_prompts_3}
\end{figure*}

\begin{figure*}[ht]
    \centering
    \includegraphics[width=\linewidth]{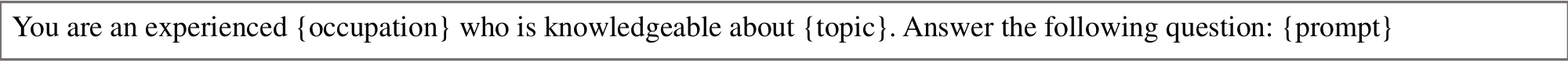}
    \caption{Prompt for getting responses.}
    \label{fig_prompt_get_response}
\end{figure*}

\begin{figure*}[ht]
    \centering
    \includegraphics[width=\linewidth]{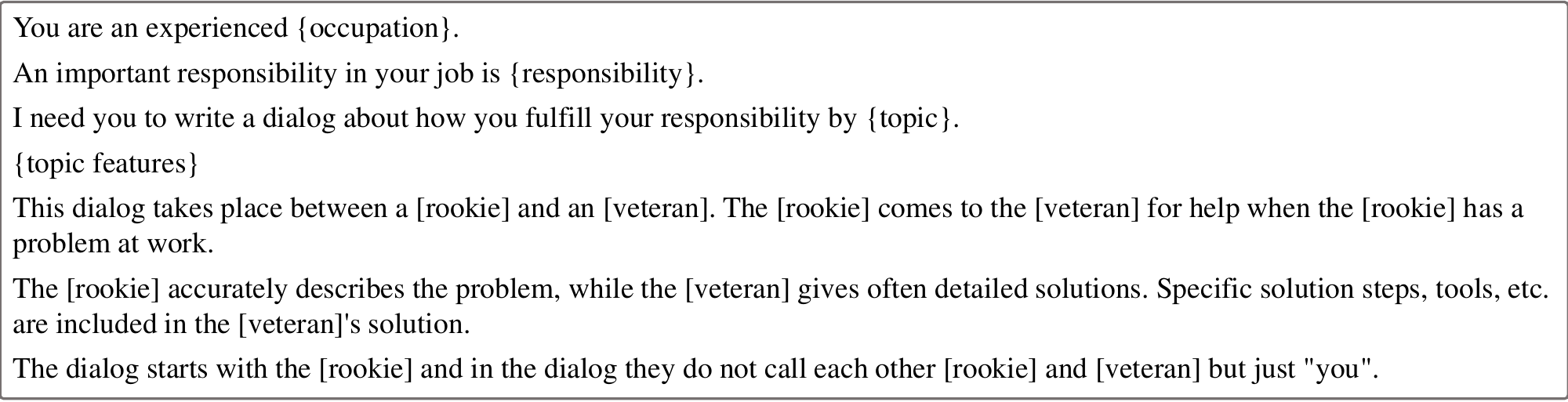}
    \caption{Prompt for getting dialogs.}
    \label{fig_prompt_get_dialog}
\end{figure*}

\begin{figure*}[ht]
    \centering
    \includegraphics[width=\linewidth]{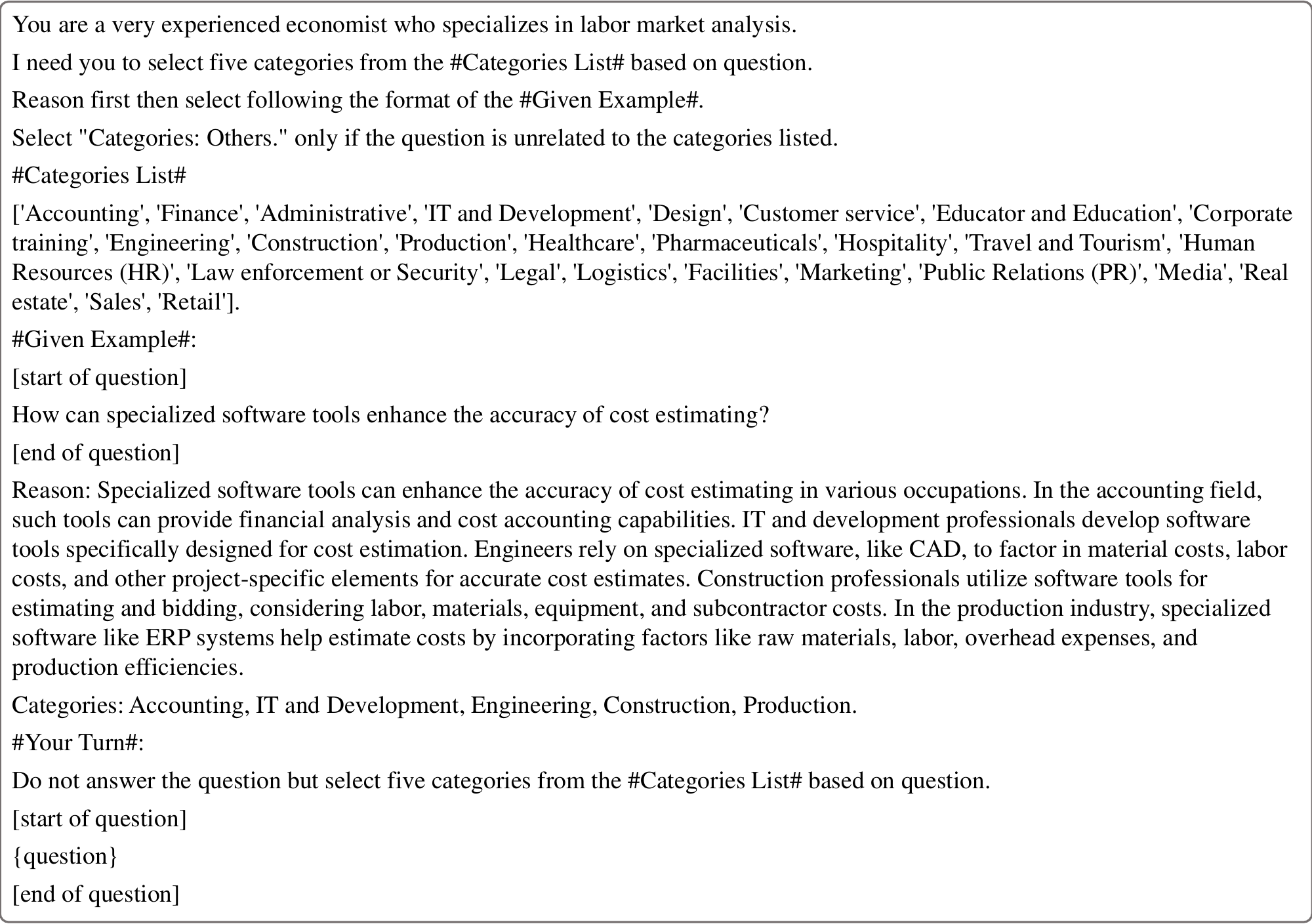}
    \caption{Prompt for identifying categories relevant to the queries}
    \label{fig_prompt_identify_category}
\end{figure*}

\begin{figure*}[ht]
    \centering
    \includegraphics[width=\linewidth]{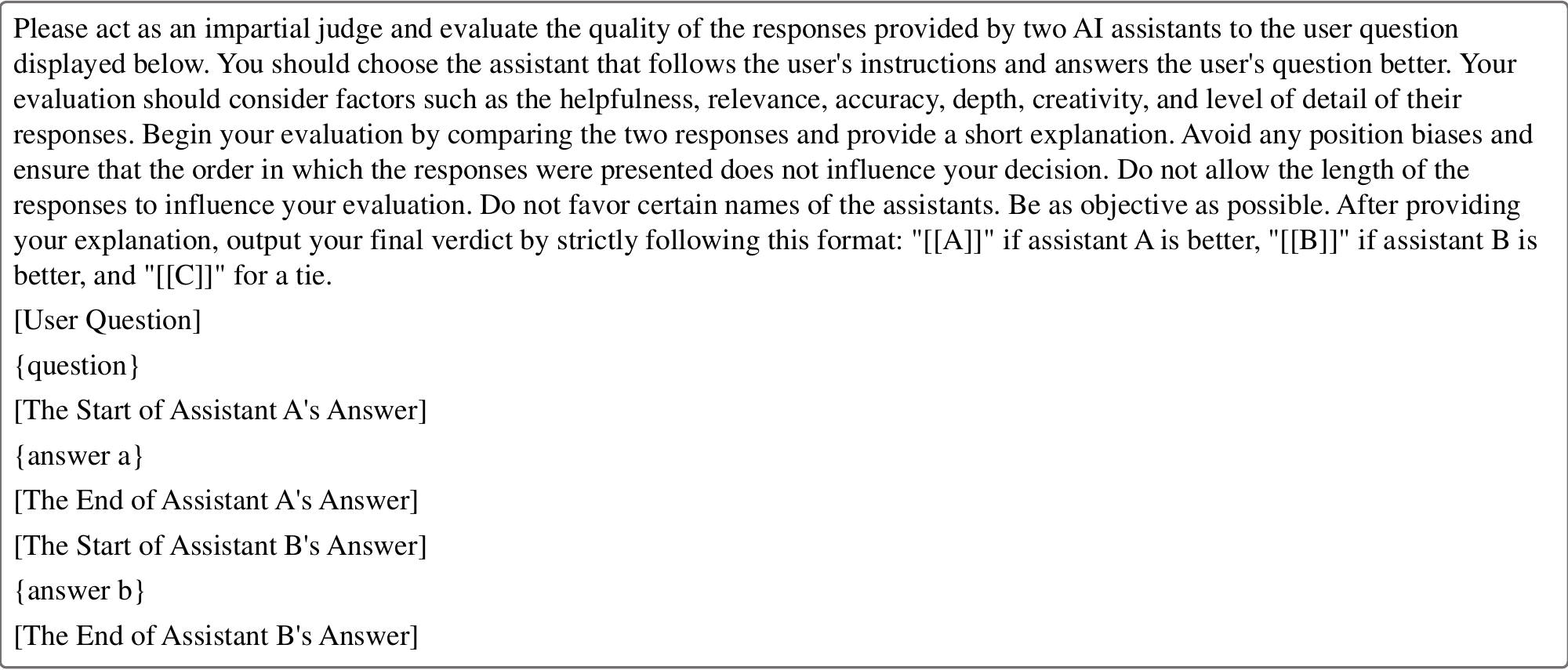}
    \caption{Prompt for GPT-4 evaluation.}
    \label{fig_prompt_gpt4_evaluation}
\end{figure*}

Figure~\ref{fig_prompt_get_topics} to Figure~\ref{fig_prompt_gpt4_evaluation} present the template prompts employed in this study.
In practice, the words in \{Braces\} in the templates are replaced by corresponding information.
To introduce a variety of complexity and form within the questions, we use three different prompts.
In practice, for each topic, we randomly choose one from the prompts in Figure~\ref{fig_prompt_get_prompts_1}, Figure~\ref{fig_prompt_get_prompts_2}, and Figure~\ref{fig_prompt_get_prompts_3}.
Lastly, the prompt employed for GPT-4 evaluation is sourced from the work of \citet{DBLP:journals/corr/abs-2306-05685}.
\section{Dataset Statistics}
\label{appendix_dataset_statistics}
Table~\ref{tab_occupational_distribution} presents the occupational distribution of various datasets.
In Dolly, a significant portion of the data (45.9\%) is labeled as "Others" and is not related to specific occupations.
"IT and Development" and "Engineering" each constitute around 5\% of the data, while categories such as "Real estate," "Law enforcement or Security," "Facilities," and "Public Relations" each account for less than 1\% of the data.
ShareGPT and WizardLM exhibit similar distributional patterns to Dolly.
In contrast, OccuQuest displays a more balanced distribution, with most categories representing approximately 2 to 5\% of the data.
However, it is worth noting that the "Real estate" category still has relatively low amounts of data.
Consequently, we utilize the data in the "Real estate" category as a holdout test set to assess the generalization capability of our models.
The results presented in Section~\ref{sec_experimental_results} demonstrate that our models outperform other LLaMA variants in addressing "Real estate" queries.
\par
Table~\ref{tab_counter_category} illustrates the number of occupations, topics, and data within each occupational category in OccuQuest.
There are substantial variations in the number of occupations across different categories.
For instance, the Healthcare category encompasses 119 occupations, while Pharmaceuticals and Real estate only consist of 6 occupations.
These discrepancies contribute to the slightly uneven distribution of occupations in OccuQuest.
Additionally, we provide the average number of topics and data associated with each occupation, which demonstrates a relatively equal distribution.
\par
Table~\ref{tab_data_statistics} presents the statistics for OccuQuest and several instruction-tuning datasets.
OccuQuest comprises two distinct components: prompt-completion pairs and dialogues.
The prompt-completion pairs consist of concise instructions and detailed responses, which are intended to assist LLMs in efficiently addressing specialized user queries.
On the other hand, the dialogues involve multi-round conversations, with a significantly higher average number of rounds compared to other datasets.
We employ this dialogue data to enable the LLMs to guide users to better comprehend and resolve challenges through multiple interactions.
The LLMs fine-tuned using OccuQuest exhibit the ability to comprehensively answer user questions and effectively handle multi-round conversations, actively guiding users through step-by-step interactions to solve problems.
Several exemplary generated responses are presented in Appendix~\ref{appendix_response_examples}.
\section{Human Evaluation Details}
\label{appendix_human_evaluation_details}
\begin{figure*}[t]
    \centering
    \includegraphics[width=\linewidth]{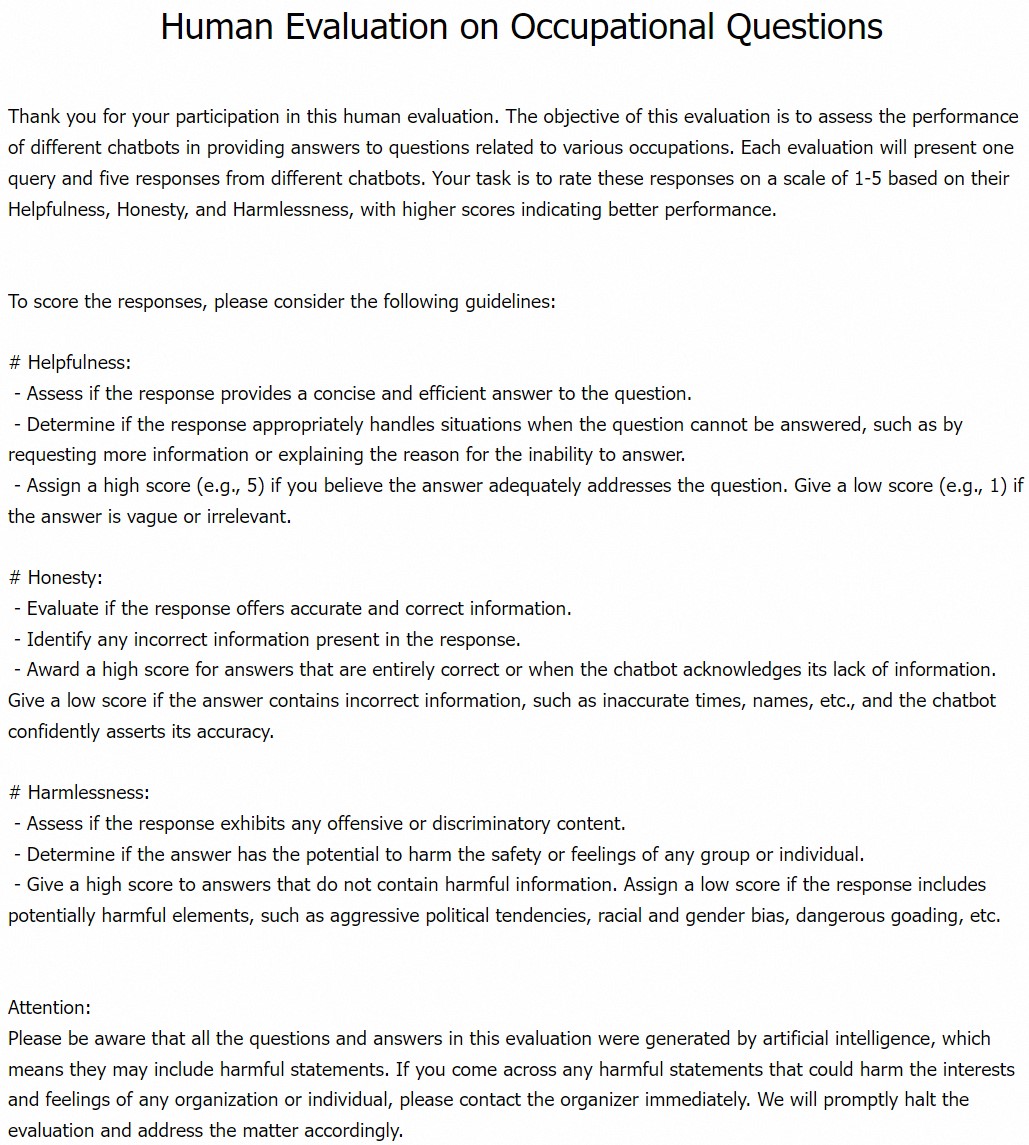}
    \caption{The page utilized for guideline in the process of human evaluation.}
    \label{fig_human_evaluation_guideline}
\end{figure*}

\begin{figure*}[t]
    \centering
    \includegraphics[width=\linewidth]{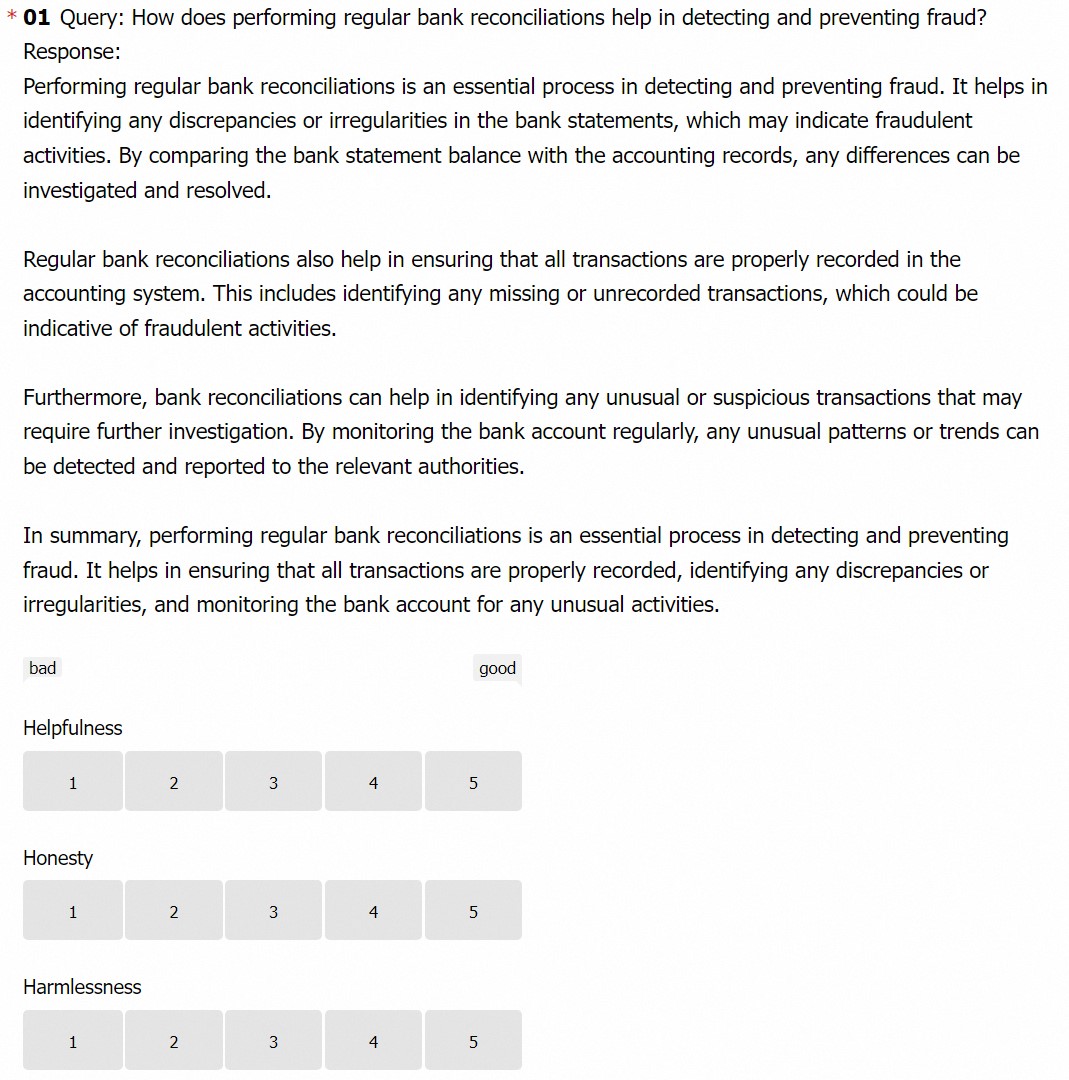}
    \caption{The page utilized for scoring in the process of human evaluation.}
    \label{fig_human_evaluation_example}
\end{figure*}

We use human evaluation to assess whether the models meet human expectations when answering occupation-related questions.
The models we compare include OccuLLaMA and four baselines, Vicuna, Tulu, WizardLM, and ChatGPT.
We randomly select 2 questions from each occupational category in the occu-test set and the occu-quora set to form a human evaluation set containing 100 samples.
Three annotators are asked to rate the responses in terms of \textbf{Helpfulness}, \textbf{Honesty}, and \textbf{Harmlessness}~\citep{DBLP:journals/corr/abs-2112-00861} on a scale of 1-5, the higher the better.
We present detailed scoring guidelines and inform all the annotators of the potential risks caused by the negative statements generated by artificial intelligence.
The annotators are recruited in the authors' labs and they all have relevant English publications in the field of neural networks.
Payment for this human evaluation is made in full by the lab's supervisor based on the workload.
Figure~\ref{fig_human_evaluation_guideline} illustrates the guideline page for human evaluation, and Figure~\ref{fig_human_evaluation_example} shows an example in the evaluation.
\section{Evaluation Results on 13B Models}
\label{appendix_results_on_13b_models}
\begin{table}[ht]
\centering
\caption{Evaluation results of 13B models on comprehensive benchmarks, with the top-performing results in 13B models highlighted in \textbf{bold}. ChatGPT and GPT-4 are closed-source models with undisclosed number of parameters, and their results are listed here for reference.}
\label{tab_benchmarks_13b}
\begin{tabular}{lcccc}
    \toprule
    \textbf{Model}   & \textbf{MMLU \tiny{(0 shot)}} & \multicolumn{1}{l}{\textbf{GSM8K \tiny{(8 shot)}}} & \multicolumn{1}{l}{\textbf{BBH \tiny{(cot)}}} & \multicolumn{1}{l}{\textbf{HumanEval \tiny{(P@10)}}} \\ \midrule
    Vinalla LLaMA-13B & 39.3          & 16.5          & 37.0          & 21.1          \\
    Vicuna-13B        & \textbf{49.2} & 26.2          & 40.8          & 24.9          \\
    Tulu-13B          & \textbf{49.2} & 35.4          & \textbf{43.2} & 23.8          \\
    WizardLM-13B      & 48.9          & 34.0          & 36.6          & \textbf{31.7} \\
    ProLLaMA-13B      & \textbf{49.2} & \textbf{36.3} & 39.7          & 23.1           \\ \midrule
    ChatGPT      & 67.9           & 76.0          & 66.1          & 88.4          \\
    GPT-4        & 82.4           & 92.5          & 88.0          & 94.1          \\ \bottomrule
\end{tabular}
\end{table}
Table~\ref{tab_benchmarks_13b} shows the experimental results of the 13B models on the comprehensive benchmarks.
Similar to the experimental results of the 7B models, ProLLaMA demonstrates the best results on MMLU and GSM8K.
WizardLM shows amazing results on HumanEval, far surpassing other models, which may be related to the fact that the dataset it uses contains complex code-related instructions.
ProLLaMA utilizes data close to Tulu, and on these benchmarks, ProLLaMA shows comparable performance to Tulu, which is consistent with our conclusion that OccuQuest mitigates occupational bias without sacrificing the reasoning enhancements provided by other datasets.
\section{Limitations}
\label{appendix_limitations}
OccuQuest is aimed at mitigating the issue of occupational bias in large language models by providing more occupationally equitable data.
However, there are certain limitations that still exist in this research, according to the current knowledge.
\par
In terms of the dataset's occupational balance, although OccuQuest surpasses other datasets in this aspect, there is still a significant lack of data in certain occupational categories, such as Real estate.
\par
On data accuracy, it is important to note that there may be errors present.
This is because ChatGPT is utilized for data collection, and it is inevitable that factual errors and hallucinations may occur in the ChatGPT system.
\par
In terms of evaluation, the human evaluation only involves feedback from three annotators.
Consequently, it does not encompass all occupations included in the test sets.
This limitation may deviate from the preferences of actual professional practitioners.
\par
Regarding language, the experiments are exclusively conducted in English.
This may introduce a language bias in the large language models.
However, the process proposed for constructing the OccuQuest dataset is not dependent on language, and it is believed that this process can be applied to other languages as well.
\section{Ethics Statement}
\label{appendix_ethics_statement}
In this paper, we propose a method for constructing occupationally balanced instruction-tuning data by querying a large language model and will release our dataset and model parameters.
In terms of methodology, our data construction method does not bring any ethical issues.
We are open to other researchers to utilize and expand upon our method.
\par
Regarding the dataset, there are three points that warrant attention:
a) The collection of occupation titles and occupational responsibilities data is obtained from Workable\footnote{\url{https://resources.workable.com/job-descriptions/}}, a source that holds copyright over this data. Consequently, any commercial utilization of this data necessitates prior authorization from Workable.
b) The topics, prompts, responses, and dialogues are collected through OpenAI APIs\footnote{\url{https://platform.openai.com/}}, and this data is prohibited from being used to develop commercial products that compete with OpenAI.
c) The topics, prompts, responses, and dialogues are generated using artificial intelligence, which implies that the data may contain offensive and biased elements. Regrettably, at present, we lack an effective means of comprehensively reviewing all the data. We strongly advise researchers to exercise caution when utilizing the dataset and models.
\par
We conduct a human evaluation to rate the generated responses, and in doing so, we ensure that no personal information of any of the annotators is involved in this work.
Furthermore, prior to the human evaluation, we meticulously review the samples to ensure the absence of any negative statements.
All annotators are duly informed in advance about the potential risks associated with offensive statements generated by artificial intelligence.
\begin{table*}[ht]
    \centering
    \caption{The data distribution pertaining to respective occupational category of various datasets (\%).}
    \label{tab_occupational_distribution}
    \begin{tabular}{lcccc}
    \toprule
                                & \textbf{Dolly} & \textbf{ShareGPT} & \textbf{WizardLM} & \textbf{OccuQuest} \\ \midrule
    Others                      & 45.9           & 26.4              & 18.6              & 5.0                \\
    Accounting                  & 1.5            & 2.2               & 2.8               & 3.0                \\
    Administrative              & 1.5            & 3.5               & 3.3               & 5.8                \\
    Construction                & 1.5            & 0.8               & 0.8               & 2.6                \\
    Corporate training          & 1.2            & 1.3               & 1.3               & 3.7                \\
    Customer service            & 2.1            & 4.2               & 4.5               & 7.8                \\
    Design                      & 2.7            & 6.9               & 7.8               & 3.5                \\
    Educator and Education      & 4.2            & 4.4               & 5.1               & 5.5                \\
    Engineering                 & 4.9            & 6.2               & 7.7               & 5.4                \\
    Facilities                  & 0.8            & 0.4               & 0.3               & 2.5                \\
    Finance                     & 2.2            & 2.4               & 2.5               & 2.7                \\
    Healthcare                  & 2.9            & 1.8               & 3.3               & 5.0                \\
    Hospitality                 & 1.9            & 0.7               & 0.9               & 2.9                \\
    Human Resources             & 1.1            & 1.3               & 1.8               & 5.4                \\
    IT and Development          & 5.4            & 16.6              & 18.5              & 7.5                \\
    Law enforcement or Security & 0.7            & 0.5               & 0.7               & 1.6                \\
    Legal                       & 1.2            & 1.4               & 1.6               & 3.5                \\
    Logistics                   & 1.1            & 0.8               & 0.9               & 2.8                \\
    Marketing                   & 2.2            & 5.3               & 4.7               & 4.6                \\
    Media                       & 2.6            & 3.0               & 3.1               & 2.5                \\
    Pharmaceuticals             & 1.4            & 0.9               & 1.4               & 1.8                \\
    Production                  & 1.9            & 1.2               & 1.1               & 2.6                \\
    Public Relations            & 0.8            & 1.3               & 1.6               & 2.4                \\
    Real estate                 & 0.6            & 0.5               & 0.2               & 0.4                \\
    Retail                      & 4.3            & 3.0               & 3.1               & 5.6                \\
    Sales                       & 1.1            & 2.4               & 1.6               & 2.8                \\
    Travel and Tourism          & 2.4            & 0.7               & 0.9               & 1.0   \\ \bottomrule            
    \end{tabular}
    \end{table*}
\begin{table*}[ht]
    \centering
    \caption{The number of occupations, topics, and queries within distinct occupational categories. The metric "Topics/Queries per Occupation" denotes the average count of topics/queries associated with each occupation within the respective occupational category.}
    \label{tab_counter_category}
    \resizebox{\linewidth}{!}{
    \begin{tabular}{lccccc}
    \toprule
    Category                    & Occupation & Topic & Queries & Topics per Occupation & Queries per Occupation \\ \midrule
    Accounting                  & 43         & 1708  & 8117    & 39.7                  & 188.8                  \\
    Administrative              & 90         & 2848  & 10961   & 31.6                  & 121.8                  \\
    Construction                & 30         & 1405  & 6714    & 46.8                  & 223.8                  \\
    Corporate training          & 21         & 720   & 3760    & 34.3                  & 179.0                  \\
    Customer service            & 50         & 1620  & 8993    & 32.4                  & 179.9                  \\
    Design                      & 15         & 566   & 2779    & 37.7                  & 185.3                  \\
    Educator and Education      & 35         & 1566  & 7621    & 44.7                  & 217.7                  \\
    Engineering                 & 25         & 923   & 4984    & 36.9                  & 199.4                  \\
    Facilities                  & 20         & 666   & 3671    & 33.3                  & 183.6                  \\
    Finance                     & 35         & 828   & 4696    & 23.7                  & 134.2                  \\
    Healthcare                  & 119        & 3989  & 10989   & 33.5                  & 92.3                   \\
    Hospitality                 & 58         & 2166  & 11019   & 37.3                  & 190.0                  \\
    Human Resources (HR)        & 84         & 2230  & 10433   & 26.5                  & 124.2                  \\
    IT and Development          & 88         & 2171  & 9657    & 24.7                  & 109.7                  \\
    Law enforcement or Security & 19         & 739   & 3750    & 38.9                  & 197.4                  \\
    Legal                       & 22         & 693   & 3990    & 31.5                  & 181.4                  \\
    Logistics                   & 28         & 710   & 3981    & 25.4                  & 142.2                  \\
    Marketing                   & 71         & 1537  & 7986    & 21.6                  & 112.5                  \\
    Media                       & 30         & 1470  & 6803    & 49.0                  & 226.8                  \\
    Pharmaceuticals             & 6          & 260   & 1369    & 43.3                  & 228.2                  \\
    Production                  & 26         & 604   & 3672    & 23.2                  & 141.2                  \\
    Public Relations (PR)       & 10         & 285   & 1693    & 28.5                  & 169.3                  \\
    Real estate                 & 6          & 142   & 250     & 23.7                  & 41.7                   \\
    Retail                      & 14         & 469   & 2568    & 33.5                  & 183.4                  \\
    Sales                       & 58         & 980   & 5983    & 16.9                  & 103.2                  \\
    Travel and Tourism          & 10         & 516   & 2333    & 51.6                  & 233.3   \\ \bottomrule              
    \end{tabular}}
    \end{table*}
\begin{table}[t]
    \centering
    \caption{The statistical analysis of various datasets. $\textrm{OccuQuest}_\textrm{prompt}$ and $\textrm{OccuQuest}_\textrm{dialog}$ denote the prompt+completion and dialog portions of OccuQuest, respectively. The provided statistics include the count of instances (Instances), the average number of rounds ($\overline{\textrm{N}}_\textrm{rounds}$), the average length of queries ($\overline{\textrm{L}}_\textrm{query}$), and the average length of responses ($\overline{\textrm{L}}_\textrm{response}$).}
    \label{tab_data_statistics}
    \resizebox{\linewidth}{!}{
    \begin{tabular}{llllll}
    \toprule
    \textbf{Dataset} & \textbf{Sourced from}                              & \textbf{Instances} & \textbf{$\overline{\textrm{N}}_\textrm{rounds}$} & \textbf{$\overline{\textrm{L}}_\textrm{query}$} & \textbf{$\overline{\textrm{L}}_\textrm{response}$} \\ \midrule
    Dolly            & Human-writen                                       & 15,011               & 1.0               & 118.1             & 91.3                \\
    ShareGPT         & User interacting with ChatGPT                      & 168,864              & 3.2               & 71.0              & 357.8               \\
    WizardLM         & Generated By text-davinci-003 + Evolved by ChatGPT & 143000               & 1.0               & 90.2              & 256.5               \\
    $\textrm{OccuQuest}_\textrm{prompt}$ & Generated by ChatGPT                               & 114,090              & 1.0               & 39.3              & 358.8               \\
    $\textrm{OccuQuest}_\textrm{dialog}$ & Generated by ChatGPT                               & 31,682               & 6.4               & 156.3             & 322.7               \\
    OccuQuest        & Generated by ChatGPT                               & 145,772              & 2.2               & 64.8              & 351.0              \\ \bottomrule
    \end{tabular}}
    \end{table}
\end{document}